\documentclass{article}
\usepackage{iclr2027_conference,times}
\usepackage{amsmath,amssymb,booktabs,graphicx,array,tabularx,multirow,xcolor,colortbl,url}
\usepackage{capt-of,needspace}
\newcounter{sketchalgorithm}
\usepackage[switch]{lineno}
\usepackage[hidelinks]{hyperref}
\usepackage{placeins}
\newif\ifprojecteddraft
\projecteddrafttrue
\definecolor{QOnepos}{HTML}{DCEBF3}
\definecolor{QOneneg}{HTML}{F9E1D5}
\definecolor{QOnePending}{HTML}{F1F3F4}
\definecolor{QHeadingColor}{HTML}{003F73}
\font\MAGboldfont = ptmb at 9.6pt
\newcommand{\BoldLabel}[1]{{\MAGboldfont #1}}

\definecolor{magBlue}{HTML}{3F6FA8}
\definecolor{magOrange}{HTML}{E6A23C}
\definecolor{magTeal}{HTML}{2A9D8F}
\definecolor{magRed}{HTML}{C95C5C}
\definecolor{magPurple}{HTML}{756BB1}
\definecolor{magGray}{HTML}{6B7280}
\definecolor{magInk}{HTML}{20242A}
\definecolor{magGuide}{HTML}{D9DEE6}
\definecolor{refBlue}{HTML}{5B9BD5}
\definecolor{refCyan}{HTML}{59C3C3}
\definecolor{refGreen}{HTML}{70C17B}
\definecolor{refYellow}{HTML}{F2CF66}
\definecolor{refOrange}{HTML}{F3A45B}
\definecolor{refCoral}{HTML}{E87A72}
\definecolor{refPurple}{HTML}{9A8BC1}
\definecolor{neighborL}{HTML}{EDF5FC}
\definecolor{automatedL}{HTML}{ECF8F4}
\definecolor{projectedL}{HTML}{FFF5E5}
\definecolor{purpleL}{HTML}{F1EEFA}
\definecolor{warningL}{HTML}{FCEDEC}
\definecolor{summaryL}{HTML}{F1F2F4}
\definecolor{metaL}{HTML}{FFF8E7}
\definecolor{magL}{HTML}{EAF2FB}

\newcommand{\QOneScoutCMRRho}{0.28}
\newcommand{\QOneScoutNCMAE}{1.63}
\newcommand{\QOneNetInfoFLPRho}{0.43}
\newcommand{\QOneWDGHLPRho}{0.41}
\newcommand{\QTwoFullNCRho}{0.42}
\newcommand{\QTwoFrozenNCRho}{0.40}
\newcommand{\QTwoUniformNCRho}{0.28}
\newcommand{\QTwoNoRefinementNCRho}{0.31}
\newcommand{\QTwoFullLPRho}{0.38}
\newcommand{\QTwoNoUtilityLPRho}{0.29}
\newcommand{\QTwoFrozenLPRho}{0.35}
\newcommand{\QTwoMinCMRRho}{0.19}
\newcommand{\QThreeScoutSaved}{23.6}
\newcommand{\QThreeScoutRet}{96.7}

\newcommand{\QThreeScoutBuild}{76.7}

\newcommand{\QThreeGLEMOSRet}{60.5}

\newcommand{\QThreeMetaGLRet}{80.2}

\newcommand{\QThreeNetInfoSavedDiff}{0.5}
\newcommand{\QThreeNetInfoSavedCILow}{-6.7}
\newcommand{\QThreeNetInfoSavedCIHigh}{7.8}

\title{No-Free-Graph: Learning When Multimodal Data Should Be Graphified}

\author{
\normalfont
Zekai Chen$^{1}$ \quad
Kai Hu$^{1}$ \quad
Yuxin Zeng$^{1}$ \quad
Xunkai Li$^{1}$ \quad
Xun Wu$^{1}$ \\
Yinlin Zhu$^{2}$ \quad
Zhengyu Wu$^{1}$ \quad
Xu Wang$^{3}$ \quad
Rong-Hua Li$^{1}$ \\[1ex]
$^{1}$Beijing Institute of Technology \\
$^{2}$Sun Yat-sen University \\
$^{3}$Shandong University \\
Zekai Chen: \texttt{zackchen02@163.com}
}

\iclrfinalcopy
\begin{document}
\raggedbottom
\maketitle

\begin{abstract}
Multimodal graph learning has recently emerged as an effective paradigm for incorporating inter-entity relationships into multimodal representations. Existing studies have made substantial progress on \emph{how to construct and optimize graphs}, but rarely consider a more fundamental question: \emph{whether additional relational structures should be introduced for a given dataset and task}. Through empirical studies across diverse datasets, tasks, and graph constructors, we reveal that graphification is not consistently beneficial: introducing relational structures can provide substantial improvements in some cases, while offering limited or even negative gains. This observation motivates a new perspective that graph construction should be treated as a selective decision based on its expected utility rather than a default preprocessing step. To address this issue, we propose \textsc{MAG-Scout}, a pre-construction graph assessment framework that estimates whether introducing graph structures is beneficial before generating the complete topology. \textsc{MAG-Scout} collects limited relational evidence, analyzes its potential task-specific contribution, and estimates the expected utility of graphification together with construction cost to make a build-or-skip decision. Extensive experiments across six multimodal datasets, three downstream tasks, and diverse graph constructors demonstrate that \textsc{MAG-Scout} effectively identifies when graph structures should be introduced, avoiding \textbf{23.6\%} of graph construction work while retaining \textbf{96.7\%} of held-out positive-gain mass under the pre-registered floor.
\end{abstract}

\section{Introduction}
\label{sec:introduction}

Multimodal data provide rich entity-level information from multiple modalities, such as textual descriptions, visual contents, and other semantic attributes. Recently, multimodal graph learning has emerged as an effective paradigm to further incorporate relationships among different entities, enabling relational reasoning beyond individual multimodal representations in applications such as recommendation, retrieval, and multimodal understanding~\citep{yan2025openmag}. By constructing connections based on semantic similarity, interactions, or domain knowledge, multimodal graphs allow models to exploit complementary information from both entity attributes and external relations. However, unlike multimodal features that are usually directly available, these inter-entity relations often need to be additionally constructed before graph-based learning. Existing studies have therefore focused on designing better graph construction strategies and graph structure learning methods to discover useful relationships and optimize graph topology for downstream tasks~\citep{franceschi2019lds,chen2020idgl,jin2020prognn}.

However, an earlier question remains largely unexplored: \emph{should additional relational structures be introduced for the current multimodal data at all?} Existing multimodal graph learning pipelines implicitly assume that enriching entity representations with graph structures will provide additional benefits once a graph can be constructed. This assumption, however, does not always hold. Although relational information can complement individual multimodal features, constructed graphs may also introduce irrelevant connections, propagate noisy signals, or provide limited additional information when existing multimodal representations already capture sufficient task-related semantics. To investigate this issue, we conduct an empirical study across diverse datasets, downstream tasks, and graph construction strategies. We find that graphification exhibits highly different outcomes under different settings: some cases obtain substantial improvements over raw multimodal predictors, whereas others achieve marginal gains or even performance degradation. These observations reveal that graph construction is not an inherently beneficial operation, but a task- and construction-dependent decision whose utility should be assessed before graphification.

This problem is not directly addressed by existing graph learning methods. Graph construction and graph structure learning approaches mainly focus on discovering meaningful relations, refining graph topology, or jointly optimizing graph structures with downstream objectives~\citep{franceschi2019lds,chen2020idgl,jin2020prognn}. Meanwhile, graph learning and model optimization methods improve prediction performance after graph structures have already been obtained~\citep{park2023metagl,park2023glemos}. Although these methods substantially advance how graphs are constructed and exploited, they assume that graphification has already been selected as an appropriate representation choice. Therefore, they cannot answer the earlier question of whether constructing additional relational structures is worthwhile for a specific dataset, task, and construction strategy. A straightforward solution is to construct the graph first and evaluate its downstream performance afterwards, but this process is inefficient because the graph construction and subsequent graph learning costs have already been incurred.

Motivated by this observation, we propose \textsc{MAG-Scout}, a pre-construction framework that estimates whether introducing graph structures is beneficial before generating the complete topology. \textsc{MAG-Scout} first performs \textbf{Candidate Relation Sampling} to collect limited relational evidence without constructing the full graph. It then conducts \textbf{Graph Utility Analysis} to assess whether the sampled relations can provide complementary information beyond existing multimodal representations. Finally, \textbf{Cost-aware Utility Prediction} estimates the expected benefit of graph construction and balances the potential gain with construction cost to make the final build-or-skip decision. In this way, \textsc{MAG-Scout} complements existing graph construction methods by determining when relational structures should be introduced rather than how they should be constructed.

\textbf{Our Contributions.}
\textbf{\textit{\underline{(1) New Perspective.}}} We identify an overlooked decision problem in multimodal graph learning: whether graph construction is beneficial before the graph is built, and show that graphification does not always provide additional benefit.
\textbf{\textit{\underline{(2) New Framework.}}} We propose \textsc{MAG-Scout}, which evaluates graph construction from limited relational evidence and predicts whether a graph is expected to improve downstream performance before full graph construction.
\textbf{\textit{\underline{(3) New Evaluation.}}}
We establish a decision-oriented evaluation protocol across six multimodal datasets, three downstream tasks, and diverse graph constructors, demonstrating that \textsc{MAG-Scout} avoids unnecessary graph construction work by \textbf{23.6\%} while retaining \textbf{96.7\%} of the held-out positive-gain mass and making reliable build-or-skip decisions.

\section{Problem Formulation}
\label{sec:problem}

We formulate graphification as a pre-construction decision problem. Given raw
multimodal data, a downstream task, and a candidate graph constructor, we aim
to determine whether introducing relational structures can provide additional
utility beyond raw multimodal representations.
Let
\begin{equation}
\mathcal{D}
=
(
\mathcal V,
\{\mathcal X^{(m)}\}_{m\in\mathcal M}
),
\quad
\mathcal G_{\mathcal C}
=
\mathcal C(\mathcal D)
=
(
\mathcal V,
\mathcal E_{\mathcal C},
\mathbf A_{\mathcal C}
),
\end{equation}
where $\mathcal V$ is the entity set, $\mathcal X^{(m)}$ denotes modality
$m$, and $\mathcal E_{\mathcal C}$ denotes constructed relations. Different
constructors may induce different graphs from the same data; therefore,
$\mathcal C$ is treated as part of the graphification condition.
For a downstream task $\mathcal T$, the utilities of raw multimodal learning
and graph-based learning are defined as:

\begin{equation}
Y_{\mathrm{raw}}
=
\mathcal Q(F_{\mathrm{raw}}(\mathcal D),\mathcal T),
\quad
Y_{\mathrm{graph}}
=
\mathcal Q(F_{\mathrm{graph}}(\mathcal G_{\mathcal C},\mathcal D),\mathcal T).
\end{equation}

The graphification value is measured by:

\begin{equation}
V^*
=
Y_{\mathrm{graph}}-Y_{\mathrm{raw}},
\quad
U^*
=
V^*-\lambda_c C(\mathcal D,\mathcal C),
\end{equation}

where $C(\mathcal D,\mathcal C)$ denotes the graphification cost. However,
$U^*$ can only be obtained after graph construction and downstream learning.
Before construction, only bounded relational evidence $\mathcal S_B$ is
available. Therefore, \textsc{MAG-Scout} aims to learn:

\begin{equation}
f_\theta(\mathcal D,\mathcal T,\mathcal C,\mathcal S_B)
\rightarrow
\widehat U,
\end{equation}

which estimates graphification utility before materializing the graph. Unlike
graph structure learning methods that optimize an existing topology,
\textsc{MAG-Scout} addresses the earlier question of whether graphification
should be invoked.

\section{Empirical Study}
\label{sec:empirical}

Before presenting \textsc{MAG-Scout}, we revisit the assumption behind existing multimodal graph learning pipelines: whether constructing a graph is consistently beneficial compared with learning directly from raw multimodal observations. We measure the relative benefit using $\Delta=Y_{\mathrm{graph}}-Y_{\mathrm{raw}}$, referred to as \emph{GraphGain}, where positive and negative values indicate whether graphification improves or degrades performance. Table~\ref{tab:q1a-graphgain-contract} reports the comparison across diverse graphification settings. The results reveal that graphification is not universally beneficial: the average GraphGain varies from $-2.91$ to $+3.72$ across different cases, showing that graph structures can provide substantial improvements in some scenarios while introducing limited or even negative effects in others. Moreover, graphification utility varies significantly with different construction strategies, indicating that its effectiveness depends on the interaction among multimodal data, downstream tasks, and graph constructors rather than the dataset alone or the final graph-model performance. Therefore, neither always constructing graphs nor always relying on raw multimodal representations is optimal. This \textbf{No-Free-Graph} phenomenon motivates \textsc{MAG-Scout}, which aims to estimate whether graphification is worthwhile before paying the cost of graph construction and downstream graph learning. Detailed experimental settings and additional empirical analyses are provided in Appendix~\ref{sec:appendix_exp_detail}.

\par\vspace{2pt}\noindent\begin{minipage}{\textwidth}
\centering
\definecolor{QOnePilotpositiveBg}{HTML}{EDF7F0}
\definecolor{QOnePilotnegativeBg}{HTML}{FDF2F0}
\definecolor{QOnePilotneutralBg}{HTML}{F8FAFC}
\definecolor{QOneSummaryBg}{HTML}{F1F5F9}
\definecolor{QOnePilotpositive}{HTML}{1B7339}
\definecolor{QOnePilotnegative}{HTML}{C53929}
\definecolor{QOnePilotneutral}{HTML}{64748B}
\newcommand{\QOnePilotGainCell}[3]{\cellcolor{QOnePilot#1Bg}\textcolor{QOnePilot#1}{#2}$_{\scriptscriptstyle\pm#3}$}
\newcommand{\QOneAvgCell}[2]{\cellcolor{QOnePilot#1Bg}\textbf{\textcolor{QOnePilot#1}{#2}}}
\captionof{table}{GraphGain relative to RawOnly across six multimodal benchmarks, in percentage points ($\Delta \equiv 0$). Cells report five-seed means with standard deviations. All nine constructors are included. The Average row is their unweighted mean on each dataset.}
\label{tab:q1a-graphgain-contract}
\scriptsize
\setlength{\tabcolsep}{3.0pt}
\renewcommand{\arraystretch}{0.78}
\resizebox{\textwidth}{!}{%
\begin{tabular}{@{}l !{\vrule width 0.5pt} cc !{\vrule width 0.5pt} cc !{\vrule width 0.5pt} cc@{}}
\toprule[1.2pt]
\BoldLabel{Constructor} & \multicolumn{2}{c!{\vrule width 0.5pt}}{\BoldLabel{Node classification (Acc)}} & \multicolumn{2}{c!{\vrule width 0.5pt}}{\BoldLabel{Link prediction (AUC)}} & \multicolumn{2}{c}{\BoldLabel{Cross-modal retrieval (R@5)}} \\
\cmidrule(lr){2-3}\cmidrule(lr){4-5}\cmidrule(lr){6-7}
 & \BoldLabel{Toys} & \BoldLabel{Grocery} & \BoldLabel{Bili Music} & \BoldLabel{DY} & \BoldLabel{QB} & \BoldLabel{Bili Cartoon} \\
\midrule[0.7pt]
\BoldLabel{kNN-GCN} & \QOnePilotGainCell{positive}{+0.58}{0.31} & \QOnePilotGainCell{positive}{+2.25}{0.30} & \QOnePilotGainCell{negative}{-4.17}{1.34} & \QOnePilotGainCell{negative}{-3.77}{0.61} & \QOnePilotGainCell{negative}{-3.65}{2.03} & \QOnePilotGainCell{negative}{-0.08}{2.54} \\
\BoldLabel{MM-kNN-GCN} & \QOnePilotGainCell{positive}{+0.92}{0.36} & \QOnePilotGainCell{positive}{+1.93}{0.08} & \QOnePilotGainCell{negative}{-3.89}{1.27} & \QOnePilotGainCell{negative}{-2.39}{0.49} & \QOnePilotGainCell{positive}{+0.33}{0.82} & \QOnePilotGainCell{negative}{-0.34}{1.34} \\
\BoldLabel{AutoG} & \QOnePilotGainCell{positive}{+4.26}{0.19} & \QOnePilotGainCell{positive}{+2.92}{0.16} & \QOnePilotGainCell{negative}{-0.87}{0.49} & \QOnePilotGainCell{negative}{-3.80}{0.62} & \QOnePilotGainCell{negative}{-2.41}{1.14} & \QOnePilotGainCell{positive}{+1.56}{0.96} \\
\BoldLabel{AutoGraph} & \QOnePilotGainCell{positive}{+7.27}{0.60} & \QOnePilotGainCell{positive}{+6.79}{0.60} & \QOnePilotGainCell{positive}{+2.02}{0.49} & \QOnePilotGainCell{negative}{-2.13}{1.41} & \QOnePilotGainCell{negative}{-3.26}{0.70} & \QOnePilotGainCell{negative}{-0.34}{1.37} \\
\BoldLabel{auGraph} & \QOnePilotGainCell{positive}{+5.11}{0.29} & \QOnePilotGainCell{positive}{+4.97}{0.27} & \QOnePilotGainCell{positive}{+1.23}{0.92} & \QOnePilotGainCell{negative}{-2.81}{1.08} & \QOnePilotGainCell{negative}{-4.31}{1.83} & \QOnePilotGainCell{negative}{-0.30}{1.26} \\
\midrule[0.5pt]
\BoldLabel{LLaTA} & \QOnePilotGainCell{positive}{+3.14}{0.33} & \QOnePilotGainCell{positive}{+3.68}{0.26} & \QOnePilotGainCell{negative}{-2.16}{0.84} & \QOnePilotGainCell{negative}{-2.64}{0.81} & \QOnePilotGainCell{negative}{-2.15}{1.28} & \QOnePilotGainCell{negative}{-0.17}{1.42} \\
\BoldLabel{GAugLLM} & \QOnePilotGainCell{positive}{+1.87}{0.28} & \QOnePilotGainCell{positive}{+2.45}{0.19} & \QOnePilotGainCell{negative}{-3.45}{1.12} & \QOnePilotGainCell{negative}{-3.51}{0.58} & \QOnePilotGainCell{negative}{-3.72}{1.54} & \QOnePilotGainCell{negative}{-0.41}{1.25} \\
\BoldLabel{LAGA} & \QOnePilotGainCell{positive}{+2.63}{0.35} & \QOnePilotGainCell{positive}{+3.12}{0.22} & \QOnePilotGainCell{positive}{+0.73}{0.68} & \QOnePilotGainCell{negative}{-2.88}{0.76} & \QOnePilotGainCell{negative}{-2.83}{0.96} & \QOnePilotGainCell{positive}{+0.58}{1.08} \\
\BoldLabel{GraphMaster} & \QOnePilotGainCell{positive}{+4.78}{0.42} & \QOnePilotGainCell{positive}{+5.36}{0.37} & \QOnePilotGainCell{positive}{+1.58}{0.53} & \QOnePilotGainCell{negative}{-2.27}{1.05} & \QOnePilotGainCell{negative}{-0.56}{1.14} & \QOnePilotGainCell{positive}{+1.04}{0.91} \\
\midrule[0.7pt]
\rowcolor{QOneSummaryBg}\BoldLabel{Average} & \QOneAvgCell{positive}{+3.40} & \QOneAvgCell{positive}{+3.72} & \QOneAvgCell{negative}{-1.00} & \QOneAvgCell{negative}{-2.91} & \QOneAvgCell{negative}{-2.51} & \QOneAvgCell{positive}{+0.17} \\
\bottomrule[1.2pt]
\end{tabular}%
}
\par\vspace{1pt}
{\scriptsize Positive cells are gains over RawOnly and negative cells are losses. The Average row equals the unweighted mean of the nine constructors above.}
\end{minipage}\par\vspace{2pt}

\section{Related Work}
\label{sec:related_work}

\textbf{Graph Construction and Structure Learning.}
Graph structure learning aims to discover or refine relational structures when
explicit topology is unavailable. Early methods such as LDS~\citep{franceschi2019lds},
IDGL~\citep{chen2020idgl}, and Pro-GNN~\citep{jin2020prognn} jointly optimize
graph structures and downstream objectives, while later approaches such as
SUBLIME~\citep{liu2022sublime} and NodeFormer~\citep{wu2022nodeformer} improve
scalable and self-supervised graph induction. Recent multimodal graph methods,
including multimodal graph benchmarks~\citep{yan2025openmag},
further incorporate multimodal information into graph representations. These
methods mainly focus on improving \emph{how to construct graphs} once
graph-based learning is considered beneficial, but do not examine whether
graphification itself is worthwhile.

\textbf{Graph Utility Assessment and Selection.}
Another line of work studies the usefulness and optimization of available
graphs. Graph property analyses investigate factors such as homophily and
neighborhood consistency~\citep{zhu2020beyond,pei2020geomgcn}, while recent
studies explore when graph structures benefit prediction, including
\emph{When Do GNNs Help}~\citep{luan2023wdgh} and
NetInfoF~\citep{lee2024netinfof}. Automated graph learning methods further
address graph model selection through architecture search and performance
prediction, including GraphNAS~\citep{gao2020graphnas}, Auto-GNN~\citep{zhou2019autognn},
AutoGL~\citep{zhang2021autogl}, MetaGL~\citep{park2023metagl}, and
GLEMOS~\citep{park2023glemos}. However, these approaches assume that graph
structures have already been constructed and focus on improving graph-based
learning rather than determining whether graph construction should be performed.

\section{Methodology}
\label{sec:method}

\subsection{Overview}

\begin{figure*}[t]
  \centering
  \includegraphics[width=\textwidth]{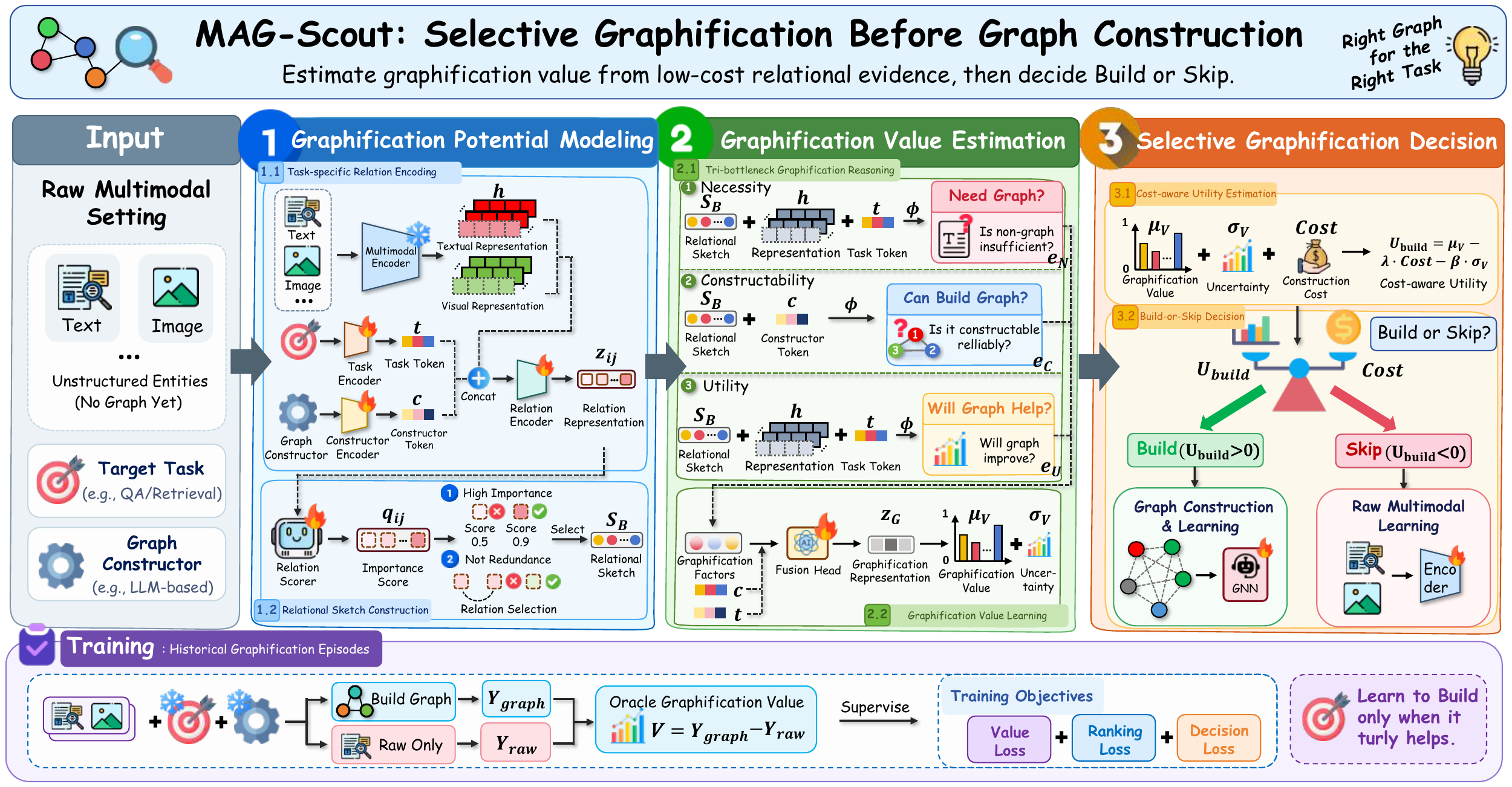}
  \caption{Overview of \textsc{MAG-Scout}: (1) Relational encoding on $\mathcal{P}$ yields compact sketch $\mathcal{S}_B$; (2) Tri-bottleneck reasoning ($\mathbf{e}_N, \mathbf{e}_C, \mathbf{e}_U$) estimates value $\mu_V$ and scale $\sigma_V$; (3) Cost-aware gating uses calibrated scale $\tilde\sigma_V$ to render Build/Skip decisions.}
  \label{fig:mag-scout-framework}
\end{figure*}

We propose \textsc{MAG-Scout}, a pre-construction framework that estimates
the value of multimodal graphification before explicitly constructing the
graph. Given multimodal data $\mathcal D$, downstream task $\mathcal T$, and
graph constructor $\mathcal C$, \textsc{MAG-Scout} learns whether introducing
relational structures can provide sufficient benefit over raw multimodal
representations under the current graphification condition. It first extracts
low-cost relational evidence, then estimates graphification value by analyzing
whether relations are necessary, constructable, and useful, and finally makes a
cost-aware build-or-skip decision. It consists of three modules:
\textbf{Graphification Potential Modeling},
\textbf{Graphification Value Estimation}, and
\textbf{Selective Graphification Decision}.
Figure~\ref{fig:mag-scout-framework} illustrates the overall architecture.

\subsection{Graphification Potential Modeling}
\label{sec:potential}

\textbf{Motivation.}
Graphification depends on relational patterns that are unavailable before
construction. We extract a compact sketch from a bounded candidate pool to
assess these patterns without materializing the complete graph.

\textbf{Task-conditioned Relational Representation.}
Given multimodal data $\mathcal D$, we obtain entity representations:

\begin{equation}
\mathbf H
=
E_{\mathrm{mm}}(\mathcal D)
=
\{\mathbf h_i\}_{i=1}^{N},
\end{equation}

where $E_{\mathrm{mm}}(\cdot)$ denotes a multimodal encoder. The task and
graph constructor are encoded as
$\mathbf t=E_{\mathrm{task}}(\mathcal T)$ and
$\mathbf c=E_{\mathrm{con}}(\mathcal C)$.
Instead of constructing all possible relations, we first obtain a bounded
candidate relation pool $\mathcal P$ through inexpensive relation discovery.
For each candidate relation $(i,j)\in\mathcal P$, we learn a
task-conditioned relational representation:

\begin{equation}
\mathbf z_{ij}
=
\phi_{\theta}
(
\mathbf h_i,
\mathbf h_j,
\mathbf h_i\odot\mathbf h_j,
|\mathbf h_i-\mathbf h_j|,
\mathbf t,
\mathbf c
),
\end{equation}

where $\mathbf z_{ij}$ describes the potential contribution of this relation
to graphification assessment rather than predicting a final graph edge.

\textbf{Relational Sketch Construction.}
Since graphification value depends on the collective properties of relations,
we select a compact relational sketch by combining relevance with a
positive-semidefinite diversity kernel. Let

\begin{equation}
\label{eq:sketch_objective}
\begin{aligned}
q_e&=\operatorname{softplus}(\mathbf u^\top\mathbf z_e),
\qquad \mathbf v_e=q_e\frac{\mathbf z_e}{\max(\|\mathbf z_e\|_2,\epsilon_z)},\\
f(\mathcal S)&=\log\det\!\left(\mathbf I_{d_z}+\sum_{e\in\mathcal S}\mathbf v_e\mathbf v_e^\top\right),
\qquad \mathcal S^\star\in\arg\max_{\mathcal S\subseteq\mathcal P,\,|\mathcal S|\le B}f(\mathcal S).
\end{aligned}
\end{equation}

Here $e=(i,j)$ indexes candidate relations, $\mathbf u$ is learned, and
$\epsilon_z>0$ guards zero embeddings. Greedy selection returns $\mathcal S_B$
with $f(\mathcal S_B)\ge(1-1/e)f(\mathcal S^\star)$ for fixed features and a
cardinality budget; Appendix~\ref{sec:appendix_sketch} gives the exact marginal and proof.

\subsection{Graphification Value Estimation}
\label{sec:value}

\textbf{Motivation.}
A sketch reveals potential relations, not their downstream benefit.
We organize value estimation around three diagnostic questions: whether raw
features lack task-relevant information, whether reliable relations can be
formed, and whether propagation sharpens the task-relevant signal.

\textbf{Tri-Bottleneck Graphification Reasoning.}
Given the sketch $\mathcal S_B$, we extract three latent states:

\begin{equation}
[
\mathbf e_N,
\mathbf e_C,
\mathbf e_U
]
=
\Phi
(
\mathcal S_B,
\mathbf H,
\mathbf t,
\mathbf c
),
\end{equation}

where $\mathbf e_N$, $\mathbf e_C$, and $\mathbf e_U$ characterize whether a
graph is needed, can be constructed, and is expected to help, respectively.
Specifically, $\mathbf e_N$ captures the insufficiency of raw multimodal
representations for the target task, $\mathbf e_C$ reflects the reliability
and coverage of candidate relations, and $\mathbf e_U$ describes the potential
benefit of relational reasoning.
The three bottlenecks are further integrated under the current task and
construction condition:

\begin{equation}
\mathbf z_G
=
f_{\psi}
(
[
\alpha_N\mathbf e_N
\Vert
\alpha_C\mathbf e_C
\Vert
\alpha_U\mathbf e_U
\Vert
\mathbf t
\Vert
\mathbf c
]
),
\quad
\alpha_k
=
\sigma(
\mathbf w_k^\top
[
\mathbf e_k\Vert\mathbf t\Vert\mathbf c
]
).
\end{equation}

Based on the resulting graphification representation, we estimate the expected
benefit and its uncertainty:

\begin{equation}
\label{eq:value_head}
(\mu_V,s_V)=F_\omega(\mathbf z_G),\quad
\sigma_V=\operatorname{softplus}(s_V)+\epsilon_\sigma,\quad
\tilde\sigma_V=a_{\mathrm{cal}}\sigma_V.
\end{equation}
Here $F_\omega$ predicts the mean and raw scale; $\sigma_V$ and $\tilde\sigma_V$
have GraphGain units. The positive calibration factor $a_{\mathrm{cal}}$ is
fitted on held-out calibration sources (Appendix~\ref{sec:appendix_training}).

\textbf{Value Learning Objective.}
During training, historical graphification episodes provide oracle values:

\begin{equation}
\label{eq:oracle_gain}
V^*
=
Y_{\mathrm{graph}}
-
Y_{\mathrm{raw}},
\end{equation}

where $Y_{\mathrm{graph}}$ and $Y_{\mathrm{raw}}$ denote downstream
performance with and without graph construction. We optimize the value
estimator with:

\begin{equation}
\label{eq:training_objective}
\mathcal L
=
\mathcal L_{\mathrm{gain}}
 +\lambda_{\mathrm{rank}}\mathcal L_{\mathrm{rank}}
 +\lambda_{\mathrm{unc}}\mathcal L_{\mathrm{unc}}
 +\lambda_{\mathrm{dec}}\mathcal L_{\mathrm{dec}},
\end{equation}

where the ranking term preserves relative graphification utility, the
uncertainty term trains the predicted spread, and the decision term aligns
value estimation with build-or-skip decisions. The uncertainty likelihood and
pairwise ranking term are defined explicitly in Appendix~\ref{sec:appendix_training}.

\subsection{Selective Graphification Decision}
\label{sec:decision}

\textbf{Motivation.}
Graphification value alone does not account for execution cost or prediction
uncertainty. The decision combines both with the estimated gain.

\textbf{Cost-aware Utility Optimization.}
Given the estimated value distribution, we formulate the utility of graph
construction as:

\begin{equation}
\label{eq:decision_score}
U_{\mathrm{build}}
=
\mu_V
-
\lambda_c \widehat C_{\mathrm{pre}}(\mathcal D,\mathcal C)
-
\lambda_u\tilde\sigma_V,
\end{equation}

Here $\widehat C_{\mathrm{pre}}$ is predicted from pre-construction features;
measured costs supervise it offline and score decisions afterward.
$\lambda_c$ converts cost into gain units; $\lambda_u\ge0$ weights calibrated uncertainty.

\textbf{Build-or-Skip Decision.}
The final action is obtained by comparing graph construction with retaining the
raw multimodal representation:

\begin{equation}
\label{eq:decision_policy}
a^*
=
\mathbb I[U_{\mathrm{build}}\geq\tau_{\mathrm{thr}}],
\end{equation}

The threshold $\tau_{\mathrm{thr}}$ is selected on inner policy-validation
sources after calibration is frozen. Action $a^*=1$ builds the graph;
$a^*=0$ retains the raw representation. Equation~\eqref{eq:decision_policy}
therefore requires neither the target graph nor its realized gain at inference.

\section{Experiments}
\label{sec:experiments}

We study four aspects of selective graphification. Q1 measures constructor-level gains and tests whether they can be predicted before graph construction. Q2 examines the effect of removing predictor components. Q3 evaluates Build/Skip policies at a 95\% retained-positive-gain target. Q4 tests sensitivity to hyperparameter changes and corrupted relational evidence.

\subsection{Experimental Setup}
\label{sec:experimental_setup}

The constructor-level comparison uses six OpenMAG benchmarks~\citep{wan2026openmag}: \textit{Toys} and \textit{Grocery} for node classification, \textit{Bili Music} and \textit{DY} for link prediction, and \textit{QB} and \textit{Bili Cartoon} for cross-modal retrieval. These are the six benchmarks; node classification, link prediction, and retrieval are the three task families. We compare RawOnly with nine graph constructors: \textbf{kNN-GCN}, \textbf{MM-kNN-GCN}, \textbf{AutoG}, \textbf{AutoGraph}, \textbf{auGraph}, \textbf{LLaTA}, \textbf{GAugLLM}, \textbf{LAGA}, and \textbf{GraphMaster}. The predictor comparison includes \textbf{\textsc{MAG-Scout}}, \textbf{NetInfoF}~\citep{lee2024netinfof}, \textbf{WDGH}~\citep{luan2023wdgh}, \textbf{GLEMOS-S2}~\citep{park2023glemos}, and \textbf{MetaGL}~\citep{park2023metagl}. Q1a uses the six benchmarks; Q1b, Q3, and Q4 use the 13-source registry; Q2 uses a separate, rebuilt 11-source bank. We report Accuracy, AUC, R@5@100, signed GraphGain, rank correlations, gain MAE, retained positive-gain mass, and saved graph work. Appendix~\ref{sec:appendix_exp_detail} gives the data splits, seed protocols, and aggregation rules.

\subsection{Is Graph Construction Always Beneficial? (Q1)}
\label{sec:q1}

Q1a measures each constructor's GraphGain relative to RawOnly on the six benchmarks. Q1b tests source-held-out prediction on the 13-source registry. Table~\ref{tab:q1a-graphgain-contract} gives the benchmark-level gains; Figure~\ref{fig:q1b-faithful-comparison} and Table~\ref{tab:q1b-pilot-scorecard} summarize predictor quality.

The nine-constructor mean ranges from $\mathbf{-2.91}$ points on DY to $\mathbf{+3.72}$ on Grocery. AutoGraph gains $\mathbf{+7.27}$ points on Toys; auGraph loses $\mathbf{4.31}$ on QB. On QB, eight constructors have negative mean gain, and all nine do on DY. Q1b asks whether information available before full graph construction predicts such differences. On the 13 parent sources, NetInfoF and WDGH reach link-prediction rank correlations of $\rho=\mathbf{\QOneNetInfoFLPRho}$ and $\rho=\mathbf{\QOneWDGHLPRho}$. \textsc{MAG-Scout} reaches $\rho=\mathbf{\QOneScoutCMRRho}$ for cross-modal retrieval and a node-classification gain MAE of $\mathbf{\QOneScoutNCMAE}$ points. All predictors receive raw multimodal features, task and constructor identifiers, and bounded relational evidence. Appendices~\ref{sec:appendix_datasets}, \ref{sec:appendix_constructors}, and \ref{sec:appendix_predictors} describe the data and models.

\par\vspace{1pt}
\begin{minipage}[t]{\textwidth}\centering
  \vspace{0pt}
  \includegraphics[width=0.327\linewidth]{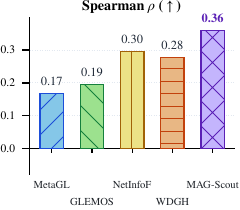}\hspace{1pt}%
  \includegraphics[width=0.327\linewidth]{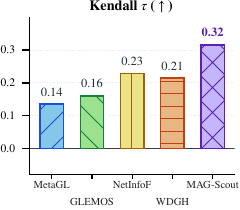}\hspace{1pt}%
  \includegraphics[width=0.327\linewidth]{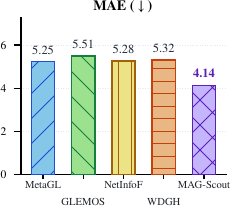}
  \captionof{figure}{Macro prediction quality across five methods and three task families on the 13 parent sources. Panels report Spearman correlation, Kendall correlation, and gain MAE.}
  \label{fig:q1b-faithful-comparison}
\end{minipage}\par\vspace{2pt}

\par\vspace{2pt}\noindent\begin{minipage}{\textwidth}
\centering
\definecolor{magL}{HTML}{EAF2FB}
\definecolor{magBlue}{HTML}{0F4D92}
\definecolor{summaryBg}{HTML}{F1F5F9}
\captionof{table}{Pre-construction prediction quality across the 13 parent sources. Cells report task-family means with seed-level standard deviations; bold and underline indicate the best and second-best values.}
\label{tab:q1b-pilot-scorecard}
\scriptsize
\setlength{\tabcolsep}{2.2pt}
\renewcommand{\arraystretch}{0.71}
\resizebox{\textwidth}{!}{%
\begin{tabular}{@{}l !{\vrule width 0.5pt} ccc !{\vrule width 0.5pt} ccc !{\vrule width 0.5pt} ccc@{}}
\toprule[1.2pt]
\BoldLabel{Predictor} & \multicolumn{3}{c!{\vrule width 0.5pt}}{\BoldLabel{Spearman rank correlation ($\rho \uparrow$)}} & \multicolumn{3}{c!{\vrule width 0.5pt}}{\BoldLabel{Kendall rank correlation ($\tau \uparrow$)}} & \multicolumn{3}{c}{\BoldLabel{Gain error ($\text{MAE} \downarrow$)}} \\
\cmidrule(lr){2-4}\cmidrule(lr){5-7}\cmidrule(lr){8-10}
 & \BoldLabel{NC} & \BoldLabel{LP} & \BoldLabel{CMR} & \BoldLabel{NC} & \BoldLabel{LP} & \BoldLabel{CMR} & \BoldLabel{NC} & \BoldLabel{LP} & \BoldLabel{CMR} \\
\midrule[0.7pt]
\rowcolor{magL}\BoldLabel{MAG-Scout} & \textbf{0.417}$_{\scriptscriptstyle\pm 0.05}$ & 0.383$_{\scriptscriptstyle\pm 0.06}$ & \textbf{0.276}$_{\scriptscriptstyle\pm 0.06}$ & \textbf{0.362}$_{\scriptscriptstyle\pm 0.03}$ & \textbf{0.331}$_{\scriptscriptstyle\pm 0.03}$ & \textbf{0.253}$_{\scriptscriptstyle\pm 0.08}$ & \textbf{1.63}$_{\scriptscriptstyle\pm 0.17}$ & \textbf{6.83}$_{\scriptscriptstyle\pm 0.45}$ & \textbf{3.97}$_{\scriptscriptstyle\pm 0.29}$ \\
\BoldLabel{NetInfoF} & \underline{0.358}$_{\scriptscriptstyle\pm 0.11}$ & \textbf{0.429}$_{\scriptscriptstyle\pm 0.08}$ & \underline{0.099}$_{\scriptscriptstyle\pm 0.13}$ & \underline{0.318}$_{\scriptscriptstyle\pm 0.07}$ & \underline{0.295}$_{\scriptscriptstyle\pm 0.06}$ & \underline{0.074}$_{\scriptscriptstyle\pm 0.06}$ & \underline{2.14}$_{\scriptscriptstyle\pm 0.24}$ & 8.48$_{\scriptscriptstyle\pm 1.53}$ & 5.22$_{\scriptscriptstyle\pm 0.53}$ \\
\BoldLabel{WDGH} & 0.336$_{\scriptscriptstyle\pm 0.12}$ & \underline{0.414}$_{\scriptscriptstyle\pm 0.07}$ & 0.079$_{\scriptscriptstyle\pm 0.12}$ & 0.294$_{\scriptscriptstyle\pm 0.11}$ & 0.282$_{\scriptscriptstyle\pm 0.06}$ & 0.068$_{\scriptscriptstyle\pm 0.08}$ & 2.39$_{\scriptscriptstyle\pm 0.22}$ & 8.74$_{\scriptscriptstyle\pm 1.36}$ & 4.82$_{\scriptscriptstyle\pm 0.79}$ \\
\BoldLabel{GLEMOS-S2} & 0.182$_{\scriptscriptstyle\pm 0.08}$ & 0.342$_{\scriptscriptstyle\pm 0.11}$ & 0.059$_{\scriptscriptstyle\pm 0.08}$ & 0.155$_{\scriptscriptstyle\pm 0.08}$ & 0.279$_{\scriptscriptstyle\pm 0.08}$ & 0.046$_{\scriptscriptstyle\pm 0.06}$ & 2.89$_{\scriptscriptstyle\pm 0.28}$ & 7.94$_{\scriptscriptstyle\pm 0.93}$ & 5.70$_{\scriptscriptstyle\pm 0.72}$ \\
\BoldLabel{MetaGL} & 0.167$_{\scriptscriptstyle\pm 0.09}$ & 0.298$_{\scriptscriptstyle\pm 0.10}$ & 0.037$_{\scriptscriptstyle\pm 0.12}$ & 0.132$_{\scriptscriptstyle\pm 0.11}$ & 0.246$_{\scriptscriptstyle\pm 0.10}$ & 0.027$_{\scriptscriptstyle\pm 0.09}$ & 3.45$_{\scriptscriptstyle\pm 0.59}$ & \underline{7.54}$_{\scriptscriptstyle\pm 1.18}$ & \underline{4.76}$_{\scriptscriptstyle\pm 0.67}$ \\
\bottomrule[1.2pt]
\end{tabular}%
}
\par\vspace{0pt}
\end{minipage}\par\vspace{2pt}

\par\vspace{1.5pt}\noindent
\subsection{Which Components Enable Reliable Assessment? (Q2)}
\label{sec:q2}

Q2 tests component contributions on a rebuilt 11-source bank, separate from the Q1 and Q3 registry. All variants use the same outer folds, three model seeds, and optimization budget. For node classification, Spearman $\rho$ is $\mathbf{\QTwoFullNCRho}$ for the full model, $\mathbf{\QTwoUniformNCRho}$ with uniform gates, and $\mathbf{\QTwoNoRefinementNCRho}$ without refinement; the frozen-sketch control reaches $\mathbf{\QTwoFrozenNCRho}$. For link prediction, the corresponding values are $\mathbf{\QTwoFullLPRho}$ for the full model, $\mathbf{\QTwoNoUtilityLPRho}$ without the utility branch, and $\mathbf{\QTwoFrozenLPRho}$ for the frozen sketch. Removing the utility branch and replacing learned gates account for the largest ranking drops. Across the ten variants, cross-modal retrieval correlations are at least $\mathbf{\QTwoMinCMRRho}$. Table~\ref{tab:q2-ablation-summary} gives the absolute results; Appendix~\ref{sec:appendix_finegrained} gives the differences from the full model.

\par\vspace{1.5pt}
\noindent
\begin{minipage}[t]{0.38\textwidth}
  \vspace{0pt}
  \centering
  \includegraphics[width=\linewidth]{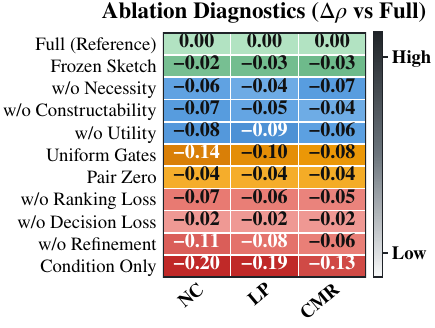}
\end{minipage}%
\hfill
\begin{minipage}[t]{0.59\textwidth}
  \vspace{0pt}
  The Full row serves as the reference. Subsequent rows isolate the three diagnostic signals ($\mathbf{e}_N, \mathbf{e}_C, \mathbf{e}_U$), followed by fusion, interaction, objective, refinement, and frozen-sketch variants.
  \par\vspace{3pt}
  \captionof{figure}{Matched ablation results relative to Full MAG-Scout on the rebuilt source bank. Colors encode the direction and magnitude of Spearman differences.}
  \label{fig:q2-ablation-heatmap}
\end{minipage}

\par\vspace{1.5pt}
\par\vspace{2pt}\noindent\begin{minipage}{\textwidth}
\centering
\definecolor{magL}{HTML}{EAF2FB}
\captionof{table}{Matched component-ablation results on the separate rebuilt source bank across node classification, link prediction, and cross-modal retrieval. Cells report means with seed-level standard deviations; bold marks the best value in each column.}
\label{tab:q2-ablation-summary}
\scriptsize
\setlength{\tabcolsep}{2.2pt}
\renewcommand{\arraystretch}{0.64}
\resizebox{\textwidth}{!}{%
\begin{tabular}{@{}l !{\vrule width 0.5pt} ccc !{\vrule width 0.5pt} ccc !{\vrule width 0.5pt} ccc@{}}
\toprule[1.2pt]
\textbf{Ablation Variant} & \multicolumn{3}{c!{\vrule width 0.5pt}}{\textbf{Spearman $\rho$ ($\uparrow$)}} & \multicolumn{3}{c!{\vrule width 0.5pt}}{\textbf{Kendall $\tau$ ($\uparrow$)}} & \multicolumn{3}{c}{\textbf{Gain error (MAE $\downarrow$)}} \\
\cmidrule(lr){2-4}\cmidrule(lr){5-7}\cmidrule(lr){8-10}
 & \textbf{NC} & \textbf{LP} & \textbf{CMR} & \textbf{NC} & \textbf{LP} & \textbf{CMR} & \textbf{NC} & \textbf{LP} & \textbf{CMR} \\
\midrule[0.7pt]
\rowcolor{magL}Full MAG-Scout (Reference) & \textbf{0.418}$_{\scriptscriptstyle\pm 0.058}$ & \textbf{0.379}$_{\scriptscriptstyle\pm 0.047}$ & \textbf{0.276}$_{\scriptscriptstyle\pm 0.071}$ & \textbf{0.361}$_{\scriptscriptstyle\pm 0.034}$ & \textbf{0.324}$_{\scriptscriptstyle\pm 0.041}$ & \textbf{0.238}$_{\scriptscriptstyle\pm 0.052}$ & \textbf{1.716}$_{\scriptscriptstyle\pm 0.112}$ & \textbf{6.748}$_{\scriptscriptstyle\pm 0.418}$ & \textbf{3.964}$_{\scriptscriptstyle\pm 0.346}$ \\
Frozen Heuristic Sketch & 0.396$_{\scriptscriptstyle\pm 0.041}$ & 0.351$_{\scriptscriptstyle\pm 0.053}$ & 0.241$_{\scriptscriptstyle\pm 0.036}$ & 0.342$_{\scriptscriptstyle\pm 0.039}$ & 0.297$_{\scriptscriptstyle\pm 0.046}$ & 0.209$_{\scriptscriptstyle\pm 0.033}$ & 1.893$_{\scriptscriptstyle\pm 0.168}$ & 7.164$_{\scriptscriptstyle\pm 0.514}$ & 4.218$_{\scriptscriptstyle\pm 0.431}$ \\
\midrule[0.5pt]
w/o Necessity ($\mathbf{e}_N$) & 0.361$_{\scriptscriptstyle\pm 0.064}$ & 0.341$_{\scriptscriptstyle\pm 0.049}$ & 0.208$_{\scriptscriptstyle\pm 0.043}$ & 0.312$_{\scriptscriptstyle\pm 0.058}$ & 0.286$_{\scriptscriptstyle\pm 0.051}$ & 0.174$_{\scriptscriptstyle\pm 0.064}$ & 2.163$_{\scriptscriptstyle\pm 0.246}$ & 7.514$_{\scriptscriptstyle\pm 0.581}$ & 4.581$_{\scriptscriptstyle\pm 0.374}$ \\
w/o Constructability ($\mathbf{e}_C$) & 0.347$_{\scriptscriptstyle\pm 0.057}$ & 0.328$_{\scriptscriptstyle\pm 0.038}$ & 0.239$_{\scriptscriptstyle\pm 0.082}$ & 0.301$_{\scriptscriptstyle\pm 0.043}$ & 0.274$_{\scriptscriptstyle\pm 0.067}$ & 0.206$_{\scriptscriptstyle\pm 0.072}$ & 2.087$_{\scriptscriptstyle\pm 0.138}$ & 7.628$_{\scriptscriptstyle\pm 0.894}$ & 4.443$_{\scriptscriptstyle\pm 0.476}$ \\
w/o Utility ($\mathbf{e}_U$) & 0.339$_{\scriptscriptstyle\pm 0.046}$ & 0.286$_{\scriptscriptstyle\pm 0.054}$ & 0.214$_{\scriptscriptstyle\pm 0.051}$ & 0.287$_{\scriptscriptstyle\pm 0.031}$ & 0.241$_{\scriptscriptstyle\pm 0.063}$ & 0.183$_{\scriptscriptstyle\pm 0.038}$ & 2.458$_{\scriptscriptstyle\pm 0.184}$ & 8.247$_{\scriptscriptstyle\pm 0.512}$ & 4.916$_{\scriptscriptstyle\pm 0.467}$ \\
\midrule[0.5pt]
Uniform Bottleneck Gates & 0.281$_{\scriptscriptstyle\pm 0.052}$ & 0.274$_{\scriptscriptstyle\pm 0.044}$ & 0.191$_{\scriptscriptstyle\pm 0.037}$ & 0.246$_{\scriptscriptstyle\pm 0.054}$ & 0.229$_{\scriptscriptstyle\pm 0.069}$ & 0.167$_{\scriptscriptstyle\pm 0.047}$ & 2.341$_{\scriptscriptstyle\pm 0.171}$ & 7.906$_{\scriptscriptstyle\pm 0.687}$ & 4.729$_{\scriptscriptstyle\pm 0.385}$ \\
w/o Pairwise Interaction & 0.381$_{\scriptscriptstyle\pm 0.079}$ & 0.342$_{\scriptscriptstyle\pm 0.036}$ & 0.237$_{\scriptscriptstyle\pm 0.062}$ & 0.328$_{\scriptscriptstyle\pm 0.028}$ & 0.291$_{\scriptscriptstyle\pm 0.042}$ & 0.204$_{\scriptscriptstyle\pm 0.045}$ & 1.924$_{\scriptscriptstyle\pm 0.119}$ & 7.237$_{\scriptscriptstyle\pm 0.642}$ & 4.306$_{\scriptscriptstyle\pm 0.428}$ \\
\midrule[0.5pt]
w/o Ranking Loss ($\mathcal{L}_{\mathrm{rank}}$) & 0.352$_{\scriptscriptstyle\pm 0.083}$ & 0.318$_{\scriptscriptstyle\pm 0.048}$ & 0.224$_{\scriptscriptstyle\pm 0.074}$ & 0.298$_{\scriptscriptstyle\pm 0.057}$ & 0.271$_{\scriptscriptstyle\pm 0.049}$ & 0.193$_{\scriptscriptstyle\pm 0.053}$ & 1.847$_{\scriptscriptstyle\pm 0.127}$ & 7.086$_{\scriptscriptstyle\pm 0.763}$ & 4.132$_{\scriptscriptstyle\pm 0.407}$ \\
w/o Decision Loss ($\mathcal{L}_{\mathrm{dec}}$) & 0.401$_{\scriptscriptstyle\pm 0.056}$ & 0.362$_{\scriptscriptstyle\pm 0.034}$ & 0.258$_{\scriptscriptstyle\pm 0.067}$ & 0.347$_{\scriptscriptstyle\pm 0.066}$ & 0.309$_{\scriptscriptstyle\pm 0.037}$ & 0.223$_{\scriptscriptstyle\pm 0.048}$ & 2.274$_{\scriptscriptstyle\pm 0.263}$ & 7.853$_{\scriptscriptstyle\pm 0.719}$ & 4.637$_{\scriptscriptstyle\pm 0.412}$ \\
w/o One-time Refinement & 0.312$_{\scriptscriptstyle\pm 0.061}$ & 0.301$_{\scriptscriptstyle\pm 0.042}$ & 0.217$_{\scriptscriptstyle\pm 0.058}$ & 0.268$_{\scriptscriptstyle\pm 0.051}$ & 0.256$_{\scriptscriptstyle\pm 0.044}$ & 0.184$_{\scriptscriptstyle\pm 0.062}$ & 2.118$_{\scriptscriptstyle\pm 0.154}$ & 7.392$_{\scriptscriptstyle\pm 0.527}$ & 4.417$_{\scriptscriptstyle\pm 0.483}$ \\
\bottomrule[1.2pt]
\end{tabular}%
}
\end{minipage}\par\vspace{2pt}

\par\vspace{1.5pt}\noindent
\subsection{Can MAG-Scout Make Cost-effective Build/Skip Decisions? (Q3)}
\label{sec:q3}

Q3 selects one threshold per predictor on inner policy-validation sources. The threshold maximizes task-macro saved graph work subject to a task-macro retained positive-gain mass of at least 95\%. Outer-test sources are scored once at that frozen threshold (Appendix~\ref{sec:appendix_metrics}). Retention averages source-level positive-gain ratios within each task, then averages the three task families. Saved graph work is computed from measured graphification costs and is the fraction of cost avoided relative to Always-Build; decisions themselves use the pre-construction cost estimate.

\par\vspace{1.5pt}
\begin{minipage}{\textwidth}\centering
  \includegraphics[width=\linewidth]{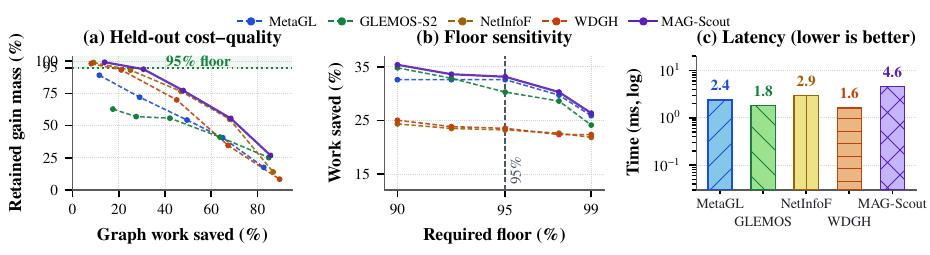}
  \captionof{figure}{Build/Skip comparison across five predictors: threshold trade-offs, retained-gain floors, and warm decision latency.}
  \label{fig:q3-system-practicality}
\end{minipage}\par\vspace{2pt}

Figure~\ref{fig:q3-system-practicality} shows the held-out policy results; Appendix~\ref{sec:appendix_operating_points} describes the operating-point metrics and Appendix~\ref{sec:appendix_floor_sweep} reports the floor sweep. At the validation-selected 95\% target, \textsc{MAG-Scout} retains $\mathbf{\QThreeScoutRet\%}$ of held-out positive-gain mass, saves $\mathbf{\QThreeScoutSaved\%}$ of graph work, and builds on $\mathbf{\QThreeScoutBuild\%}$ of cases. NetInfoF is the only other learned predictor to exceed 95\% held-out retention. The paired source-bootstrap point estimate of their saved-work difference is $\mathbf{\QThreeNetInfoSavedDiff}$ points (95\% CI $[\QThreeNetInfoSavedCILow,\ \QThreeNetInfoSavedCIHigh]$), so the observed difference is uncertain; this paired estimate need not equal the subtraction of the rounded headline values. Their wrong-skip rates are 13.3\% for \textsc{MAG-Scout} and 9.3\% for NetInfoF. WDGH's wrong-skip rate is 10.1\%, with 90.5\% retained mass. MetaGL and GLEMOS-S2 save more work, but retain only $\QThreeMetaGLRet\%$ and $\QThreeGLEMOSRet\%$; their wrong-skip rates are 27.7\% and 25.5\%. The additional savings therefore come with lower retained mass and more positive-gain cases skipped on the held-out sources.

\subsection{Is the Policy Robust to Hyperparameter and Evidence Perturbations? (Q4)}
\label{sec:q4}

Q4 uses two protocols. The sensitivity grid fixes the decision threshold at zero and varies one hyperparameter at a time. In the corruption test, we freeze the predictor and threshold selected on clean validation data, perturb the relational evidence, and measure the resulting policy changes.

\par\vspace{1.5pt}
\begin{minipage}{\textwidth}\centering
  \includegraphics[width=\linewidth]{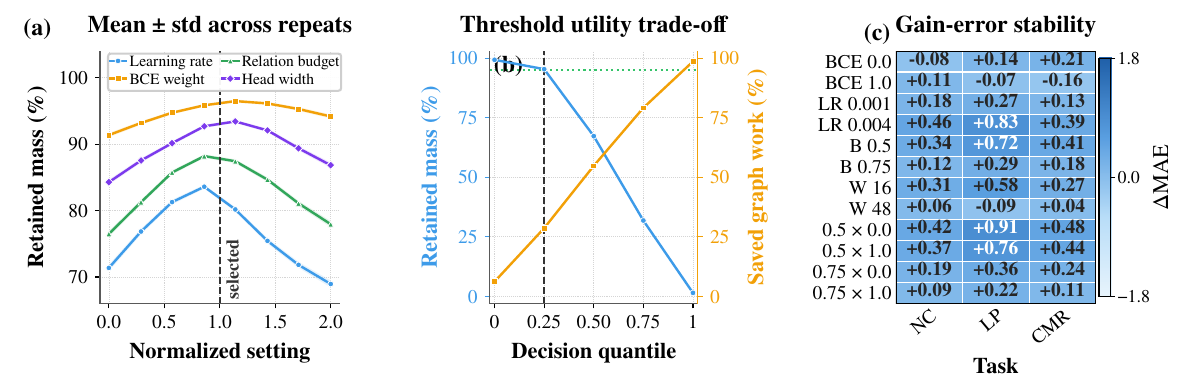}
  \captionof{figure}{One-factor retained mass at a zero threshold, the threshold trade-off, and gain error under input perturbations.}
  \label{fig:q4-stability-combined}
\end{minipage}\par\vspace{2pt}
\par\vspace{1pt}\noindent\begin{minipage}{\textwidth}
\centering
\definecolor{magL}{HTML}{EAF2FB}
\captionof{table}{Zero-threshold sensitivity (left) and frozen-policy evidence corruption (right) on 13 parent sources. Each factor has its own reference. Best (Spread) is the grid maximum and its difference from that reference.}
\label{tab:q4-stability-summary}
\scriptsize
\setlength{\tabcolsep}{3.0pt}
\renewcommand{\arraystretch}{0.65}
\resizebox{\textwidth}{!}{%
\begin{tabular}{@{}l c c c !{\vrule width 0.5pt} l c c c@{}}
\toprule[1.2pt]
\textbf{Dimension} & \textbf{Pts} & \textbf{$t=0$ Ref.} & \textbf{$t=0$ Best (Spread)} & \textbf{Perturbation} & \textbf{$\Delta$Ret.} & \textbf{$\Delta$Build} & \textbf{Flips} \\
\midrule[0.7pt]
\rowcolor{magL}Learning Rate & 3 & 81.9 & 83.6 (1.7) & Rewiring & $-0.4$ & $-7.1$ & 2/13 \\
BCE Loss Weight & 3 & 96.2 & 96.5 (0.3) & Relation Collapse & $-0.8$ & $-7.1$ & 2/13 \\
Relation Budget & 3 & 87.8 & 88.2 (0.4) & Single-Modal Dominant & $-1.6$ & $-14.3$ & 3/13 \\
Utility Head Width & 3 & 93.0 & 93.4 (0.4) & Relation-Bank Partition & $-0.1$ & $-0.2$ & 0/13 \\
\bottomrule[1.2pt]
\end{tabular}%
}
\end{minipage}\par\vspace{1pt}

\par\vspace{1.5pt}\noindent
Figure~\ref{fig:q4-stability-combined} and Table~\ref{tab:q4-stability-summary} show the two sets of results. The zero-threshold reference retention values for learning rate, BCE weight, relation budget, and utility-head width are $81.9\%$, $96.2\%$, $87.8\%$, and $93.0\%$, respectively; the largest one-factor increase is $1.7$ points. Under evidence corruption, retained mass decreases by at most $1.6$ points, while the build rate decreases by as much as $14.3$ points. At most $3/13$ parent-source decisions change; relation-bank partition changes none.

\section{Conclusion}
\label{sec:conclusion}

In this work, we revisit a fundamental yet underexplored question in multimodal graph learning: whether a graph should be constructed at all. Our empirical study reveals a \textbf{No-Free-Graph} phenomenon, showing that graphification can bring substantial gains in some cases while providing limited or even negative benefits in others. Motivated by this observation, we propose \textsc{MAG-Scout}, a pre-construction framework that estimates graphification value before the complete topology is materialized. \textsc{MAG-Scout} first extracts a compact relational sketch, then evaluates graphification from the perspectives of necessity, constructability, and utility, and finally combines the predicted value with construction cost and uncertainty to make a selective Build-or-Skip decision. Extensive experiments across diverse multimodal datasets, tasks, and graph constructors demonstrate that \textsc{MAG-Scout} can reliably identify when graphification is beneficial while avoiding unnecessary graph construction. We hope this study encourages multimodal graph learning to move beyond the assumption of always constructing graphs toward more selective and value-aware graphification.

\subsubsection*{Reproducibility and AI Use Statements}

An anonymized implementation, configurations, and reproducibility checks are available at \url{https://anonymous.4open.science/r/MAG-Scout-Review-2DC3/}. The main paper details the diagnostic protocols, the \textsc{MAG-Scout} sketch, tri-bottleneck parameterization, joint optimization, and selective decision policy; the appendix adds theoretical and robustness analyses. The code omits raw data and pretrained checkpoints but documents data format, dependencies, and reproduction instructions.

Generative AI tools assisted with language refinement, literature discovery, and research ideation. The authors reviewed and edited the assisted material and take responsibility for the manuscript.

\clearpage

\renewcommand{\refname}{References}
\bibliographystyle{iclr2027_conference}
\bibliography{references}

\clearpage

\appendix
\section{Reproducibility, Software Environment, and Hardware Infrastructure}
\label{sec:appendix_reproducibility}

\subsection{Computing Infrastructure}
The pilot evaluations ran on Linux computing servers with the following configuration.
\noindent\textbf{Processors.} Dual-socket Intel(R) Xeon(R) Gold 6240 CPUs @ 2.60\,GHz (18 physical cores per socket, 36 physical cores total, supporting 72 execution threads).
\noindent\textbf{System Memory.} 256\,GB DDR4-2933 ECC Registered RAM (with over 210\,GB dynamically available during peak distributed evaluations).
\noindent\textbf{Accelerators.} 4$\times$ NVIDIA A100 Tensor Core GPUs with 40\,GB HBM2 memory each (interconnected via PCIe Gen4 with peer-to-peer access).
\noindent\textbf{Software Stack.} Ubuntu 22.04 LTS (kernel 5.15.0-107-generic), Python 3.10.12, PyTorch 2.4.0 with CUDA 12.1 runtime and cuDNN 8.9.2, PyTorch Geometric~\citep{fey2019pyg} (\texttt{torch\_geometric}) 2.5.3, NetworkX 3.2.1, Scikit-learn 1.4.2, SciPy 1.13.0, and Matplotlib 3.8.4.

\subsection{Seed Protocol and Scientific Determinism}
Q1a, Q3, and Q4 use five matched random seeds across datasets and constructors:
\begin{equation}
    \mathcal{S}_{\mathrm{seed}} = \{344701, \, 675949, \, 86547, \, 931443, \, 798884\}.
\end{equation}
For each episode, all PyTorch, NumPy, Python standard library, and CUDA random number generator states are explicitly initialized with the corresponding seed. All bootstrap estimations (e.g., 10,000 paired bootstrap draws for saved work and retained gain mass confidence intervals) are conducted under fixed random generator states (seed \texttt{20260916}).
The current Q1b pilot package contains five runs per source with source-matched
seed sets; ten sources use the set above and three historical sources use a
source-specific five-seed set. Q2 uses three model seeds $\{17,23,42\}$ on its
separate 11-source bank. These seed scopes are reported with their respective
tables.

\subsection{Code and Asset Availability}
The accompanying paper archive contains the LaTeX sources, figure-generation scripts,
and tabulated inputs for the reported plots.

\section{Extended Methodology and Mathematical Details}
\label{sec:appendix_method_detail}

Appendix B gives implementation details for the method in Section~\ref{sec:method}, including regularized log-determinant selection, the three diagnostic heads, the four training terms, scale calibration, and threshold selection. Fitting, calibration, policy validation, and outer testing use separate source groups; Appendix~\ref{sec:appendix_training} specifies their data flow and the training objective.

\subsection{Notation}
\label{sec:appendix_notation}

\begin{table}[!htbp]
\centering
\small
\caption{Notation used throughout the methodology. Dimensions are for a single source; the encoder width is $d$ and the reduced projection width is $d' \ll d$.}
\label{tab:appendix_notation}
\setlength{\tabcolsep}{4pt}
\renewcommand{\arraystretch}{1.08}
{\footnotesize
\begin{tabular}{@{}l p{0.66\textwidth}@{}}
\toprule
\textbf{Symbol} & \textbf{Meaning} \\
\midrule
$\mathcal{D}=(\mathcal{V},\{\mathcal{X}^{(m)}\}_{m\in\mathcal{M}})$ & multimodal dataset: entity set $\mathcal{V}$ (cardinality $N=|\mathcal{V}|$) with modality tensors $\mathcal{X}^{(m)}$ \\
$\mathcal{M}$ & set of observed modalities (text, visual, and auxiliary attributes) \\
$\mathcal{T}$, $\mathcal{C}$ & downstream task and candidate graph constructor \\
$\mathcal{C}(\mathcal{D})=(\mathcal{V},\mathcal{E}_{\mathcal{C}},\mathbf{A}_{\mathcal{C}})$ & graph induced by constructor $\mathcal{C}$ \\
$\mathbf{H}=\{\mathbf{h}_i\}_{i=1}^{N}\in\mathbb{R}^{N\times d}$ & frozen multimodal entity representations ($E_{\mathrm{mm}}$) \\
$\mathbf{h}_i^{(m)}$ & modality-$m$ representation of entity $i$ \\
$\mathbf{t}=E_{\mathrm{task}}(\mathcal{T})$, $\mathbf{c}=E_{\mathrm{con}}(\mathcal{C})$ & task and constructor conditioning tokens in $\mathbb{R}^{d}$ \\
$\mathcal{P}$, $K$ & bounded candidate relation pool ($|\mathcal{P}|\le K$) and its budget \\
$\mathbf{z}_{ij}\in\mathbb{R}^{d_z}$ & task-conditioned representation of candidate relation $(i,j)$ \\
$q_{ij}=\operatorname{softplus}(\mathbf{u}^{\top}\mathbf{z}_{ij})$ & positive relevance quality of relation $(i,j)$ \\
$\mathbf v_e,\mathbf L,\mathbf M_{\mathcal S}$ & quality-weighted relation vector, Gram kernel, and $\mathbf I+\sum_{e\in\mathcal S}\mathbf v_e\mathbf v_e^\top$ \\
$\mathcal{S}_B\subseteq\mathcal P$, $B$ & greedily selected relation indices and budget; embeddings are attached to these indices \\
$\epsilon_z,\epsilon_\sigma,\epsilon_a$ & positive guards for embedding norms, predicted scales, and scale calibration \\
$\mathbf{e}_N,\mathbf{e}_C,\mathbf{e}_U\in\mathbb R^{d_e}$ & necessity, constructability, and utility latent states; $d_e$ is their width \\
$\alpha_k$ & learned gate weight for branch $k\in\{N,C,U\}$ \\
$\mathbf{z}_G$ & fused graphification representation \\
$\mu_V, s_V, \sigma_V$ & predicted mean, unconstrained scale, and positive Gaussian working-model scale \\
$a_{\mathrm{cal}},\tilde\sigma_V$ & positive calibration multiplier and calibrated scale $a_{\mathrm{cal}}\sigma_V$ \\
$V^*=Y_{\mathrm{graph}}-Y_{\mathrm{raw}}$ & realized (oracle) graphification value in percentage points \\
$\widehat C_{\mathrm{pre}}$, $C(\mathcal{D},\mathcal{C})$ & pre-construction cost estimate and measured post-hoc graphification cost \\
$U_{\mathrm{build}}$ & cost- and uncertainty-adjusted utility of building \\
$\lambda_{\mathrm{rank}},\lambda_{\mathrm{unc}},\lambda_{\mathrm{dec}},\lambda_c,\lambda_u$ & loss and utility balancing weights \\
$\pi$, $\tau_{\mathrm{thr}}$ & Build/Skip policy and its decision threshold \\
$\tau_{\mathrm{rank}},T_{\mathrm{dec}}$ & positive ranking and decision temperatures in gain units \\
\bottomrule
\end{tabular}}
\end{table}

\subsection{Candidate Relation Probing and Sketch Selection Algorithm}
\label{sec:appendix_sketch}
Section~\ref{sec:method} defines the compact sketch $\mathcal{S}_B$ selected from candidate pool $\mathcal{P}$. Algorithm~\ref{alg:relational_sketch} summarizes candidate discovery and sketch extraction.

\noindent\textbf{Low-dimensional candidate discovery.} The first stage performs approximate neighbour search in a randomly projected space. Writing $\mathbf{W}_{\mathrm{proj}}\in\mathbb{R}^{d'\times d}$ for the projection matrix with $d'\ll d$, the projected representation is
\begin{equation}
    \tilde{\mathbf{h}}_i = \frac{\mathbf{W}_{\mathrm{proj}}\mathbf{h}_i}{\max(\|\mathbf{W}_{\mathrm{proj}}\mathbf{h}_i\|_2,\epsilon_z)}\in\mathbb{R}^{d'},
\end{equation}
and cosine distance is evaluated on $\tilde{\mathbf h}_i$. Random projection
reduces the search dimension~\citep{achlioptas2003projections}, and angular
search admits locality-sensitive hashing~\citep{charikar2002simhash}.
Approximation can change neighbour order; the selection guarantee below is
relative to the resulting candidate pool, not to all $N(N-1)/2$ pairs.

\noindent\textbf{Pairwise relational representation.} For each candidate pair $(i,j)\in\mathcal{P}$ we form a feature by concatenating the two endpoint embeddings, their Hadamard product, their absolute difference, and the two conditioning tokens,
\begin{equation}
    \mathbf{z}_{ij} = \phi_{\theta}\!\big([\,\mathbf{h}_i \,\|\, \mathbf{h}_j \,\|\, \mathbf{h}_i \odot \mathbf{h}_j \,\|\, |\mathbf{h}_i - \mathbf{h}_j| \,\|\, \mathbf{t} \,\|\, \mathbf{c}\,]\big),
\end{equation}
where $\phi_{\theta}$ is a shared MLP with parameters $\theta$, $\|$ denotes concatenation, and $\odot$ the element-wise product. The input dimension is $4d+2d=6d$ and the output dimension is $d_z$. The Hadamard and difference terms expose multiplicative and contrastive interactions between the endpoints, while appending $\mathbf{t}$ and $\mathbf{c}$ conditions the representation on the current task and constructor so that the same pair may be scored differently under different graphification conditions.

\noindent\textbf{Regularized volume selection.} For $e=(i,j)\in\mathcal P$,
let $q_e=\operatorname{softplus}(\mathbf u^\top\mathbf z_e)$ and
$\mathbf v_e=q_e\mathbf z_e/\max(\|\mathbf z_e\|_2,\epsilon_z)$,
where $\mathbf u\in\mathbb R^{d_z}$ is learned and $\epsilon_z>0$.
Stacking the vectors as rows of $\mathbf V$ gives a PSD quality--diversity
kernel $\mathbf L=\mathbf V\mathbf V^\top$. The determinant identity yields
\begin{equation}
\label{eq:app_sketch_objective}
\begin{aligned}
 \mathbf M_{\mathcal S}&=\mathbf I_{d_z}+\sum_{e\in\mathcal S}\mathbf v_e\mathbf v_e^\top,\\
 f(\mathcal S)&=\log\det\mathbf M_{\mathcal S}
              =\log\det(\mathbf I_{|\mathcal S|}+\mathbf L_{\mathcal S}).
\end{aligned}
\end{equation}
This regularized volume criterion is inspired by determinant-based diversity
models~\citep{kulesza2012dpp}, but is distinct from unregularized DPP MAP
selection of $\log\det\mathbf L_{\mathcal S}$. The identity term defines
$f(\varnothing)=0$ and makes every $\mathbf M_{\mathcal S}$ positive definite,
including for duplicate or zero relation vectors. Equation~\eqref{eq:sketch_objective}
uses the same objective; $\mathcal S^\star$ denotes its optimum and
$\mathcal S_B$ the greedy approximation returned by Algorithm~\ref{alg:relational_sketch}.

\noindent\textbf{Exact incremental rule.} The matrix determinant lemma gives
\begin{equation}
\label{eq:sketch_marginal}
 \Delta(e\mid\mathcal S)=f(\mathcal S\cup\{e\})-f(\mathcal S)
 =\log\!\left(1+\mathbf v_e^\top\mathbf M_{\mathcal S}^{-1}\mathbf v_e\right).
\end{equation}
For the empty set this equals $\log(1+\|\mathbf v_e\|_2^2)$.
Candidates aligned with already selected directions receive smaller incremental
value. With $\mathbf P=\mathbf M_{\mathcal S}^{-1}$, adding $e$ permits the
rank-one update
\begin{equation}
\label{eq:sketch_update}
 \mathbf P\leftarrow\mathbf P-
 \frac{\mathbf P\mathbf v_e\mathbf v_e^\top\mathbf P}{1+\mathbf v_e^\top\mathbf P\mathbf v_e}.
\end{equation}
Cholesky solves may be used instead of explicitly forming the inverse.
For $m=|\mathcal P|$, direct marginal scanning costs $O(Bm d_z^2)$ and
stores $O(m d_z+d_z^2)$ numbers; no $m\times m$ matrix is required.

\noindent\textbf{Proposition (budgeted sketch approximation).}
For fixed finite $\{\mathbf v_e:e\in\mathcal P\}$ and integer $1\le B\le m$,
the objective in Equation~\eqref{eq:app_sketch_objective} is normalized,
monotone, and submodular. Greedy maximization of
Equation~\eqref{eq:sketch_marginal} for $B$ steps satisfies
\begin{equation}
\label{eq:sketch_bound}
 f(\mathcal S_B)\ge\left[1-\left(1-\frac1B\right)^B\right]f(\mathcal S^\star)
 \ge(1-1/e)f(\mathcal S^\star).
\end{equation}
\textit{Proof.} Normalization follows from $\det\mathbf I=1$ and
monotonicity from the nonnegative quadratic form in
Equation~\eqref{eq:sketch_marginal}. If $\mathcal S\subseteq\mathcal T$,
then $\mathbf M_{\mathcal T}\succeq\mathbf M_{\mathcal S}\succ0$, hence
$\mathbf M_{\mathcal T}^{-1}\preceq\mathbf M_{\mathcal S}^{-1}$.
The marginal therefore decreases as the selected set grows, proving
submodularity. At step $b$, submodularity and monotonicity imply that
$f(\mathcal S^\star)-f(\mathcal S_b)$ is at most the sum of the marginals
of at most $B$ elements of $\mathcal S^\star\setminus\mathcal S_b$.
Greedy captures at least $1/B$ of this gap. Iterating the gap recurrence
from $f(\varnothing)=0$ proves the bound. $\square$
The bound concerns sketch volume on the fixed candidate pool, not downstream
GraphGain; changing embeddings during selection or adding other constraints
requires a separate analysis.

\noindent\textbf{Connection to pairwise redundancy.}
For $\mathbf X=\sum_{e\in\mathcal S}\mathbf v_e\mathbf v_e^\top$, expansion in
its nonnegative eigenvalues gives
\begin{equation}
\label{eq:sketch_pairwise}
 f(\mathcal S)=\sum_e\|\mathbf v_e\|_2^2
 -\frac12\sum_{e,e'}(\mathbf v_e^\top\mathbf v_{e'})^2+R,
 \qquad 0\le R\le\frac13\operatorname{Tr}(\mathbf X^3),
\end{equation}
where both sums are over $\mathcal S$. The remainder bound follows from
$\log(1+x)=x-x^2/2+\int_0^x t^2/(1+t)\,dt$ for $x\ge0$.
Thus the criterion contains a quality term and a squared-overlap
penalty as its second-order approximation, with higher-order redundancy
handled by the determinant. The second-order expression approximates the
log-determinant objective of Equation~\eqref{eq:sketch_objective}.

\subsection{Tri-Bottleneck Latent State Formulations}
\label{sec:appendix_tribottleneck}
The tri-bottleneck block maps the sketch $\mathcal{S}_B$, entity representations $\mathbf{H}$, and conditioning tokens $\mathbf{t}, \mathbf{c}$ to three diagnostic states:
\begin{equation}
    [\mathbf{e}_N, \mathbf{e}_C, \mathbf{e}_U] = \Phi(\mathcal{S}_B, \mathbf{H}, \mathbf{t}, \mathbf{c}),
\end{equation}
Here, $\Phi$ is a shared trunk with three branch-specific heads, and each $\mathbf{e}_k\in\mathbb{R}^{d_e}$. The following subsections define the input signals and interpretation of each state.

\paragraph{1. Relational Necessity ($\mathbf{e}_N$).}
The necessity branch estimates whether the downstream task requires information beyond individual multimodal representations.
To quantify the insufficiency and ambiguity of raw features, we extract two empirical signals.
\textbf{Prediction Ambiguity ($\mathbf{x}_{\mathrm{amb}}\in\mathbb{R}^3$).} A linear probe is fit on outer-training sources and kept fixed when applied to each held-out source. Its predictions on raw multimodal features $\mathbf{H}$ yield an entropy distribution across entities:
\begin{equation}
    \mathbf{x}_{\mathrm{amb}} = \left[ \frac{1}{N}\sum_{i=1}^{N} \mathcal{H}(\hat{\mathbf{p}}_i), \; \operatorname{Var}_{i}(\mathcal{H}(\hat{\mathbf{p}}_i)), \; \frac1N\sum_{i=1}^N\operatorname{Margin}_{i}(\hat{\mathbf{p}}_i) \right],
\end{equation}
where $\hat{\mathbf{p}}_i$ is the probe's predictive distribution for entity $i$, $\mathcal{H}(\hat{\mathbf{p}}_i) = -\sum_{k} \hat{p}_{ik} \log \hat{p}_{ik}$ is the Shannon entropy~\citep{shannon1948communication} (in nats), $\operatorname{Var}_i$ is the population variance of the per-entity entropies, and $\operatorname{Margin}_i = \hat{p}_{i,(1)} - \hat{p}_{i,(2)}$ is the probability margin between the top two predicted classes with $\hat{p}_{i,(1)}\ge\hat{p}_{i,(2)}$. These statistics provide diagnostic measures of ambiguity and confidence.

\textbf{Cross-Modal Divergence ($\mathbf{x}_{\mathrm{div}}\in\mathbb{R}$).} For multimodal entities with modalities $m_1, m_2 \in \mathcal{M}$, we compute the mean cosine divergence between modality-specific representations:
\begin{equation}
    \mathbf{x}_{\mathrm{div}} = \frac{1}{N} \sum_{i=1}^{N} \left( 1 - \frac{\mathbf{h}_i^{(m_1)\top} \mathbf{h}_i^{(m_2)}}{\|\mathbf{h}_i^{(m_1)}\|_2 \|\mathbf{h}_i^{(m_2)}\|_2} \right),
\end{equation}
where the term inside the parentheses lies in $[0,2]$ and is zero when the two modality views of an entity agree. Large divergence indicates that individual modalities carry complementary (potentially contradictory) evidence, which relational structure may help reconcile. The necessity state is then parameterized as
\begin{equation}
    \mathbf{e}_N = \operatorname{MLP}_N\!\left([\,\mathbf{x}_{\mathrm{amb}} \,\|\, \mathbf{x}_{\mathrm{div}} \,\|\, \mathbf{t}\,]\right),
\end{equation}
where $\operatorname{MLP}_N$ maps the $3+1+d$ dimensional input to $d_e$.

\paragraph{2. Relational Constructability ($\mathbf{e}_C$).}
The constructability branch estimates whether the candidate constructor can build a structurally sound and reliable graph from available entities.
We quantify structural viability via two signals.
\textbf{Candidate Stability under Perturbation ($\mathbf{x}_{\mathrm{stab}}\in\mathbb{R}$).} We inject Gaussian noise $\boldsymbol{\epsilon} \sim \mathcal{N}(\mathbf{0}, 0.05^2 \mathbf{I})$ into $\mathbf{H}$, obtaining $\mathbf{H}_{\boldsymbol{\epsilon}}=\mathbf{H}+\boldsymbol{\epsilon}$, and compute the Jaccard similarity between the top-$k$ nearest neighbour candidate sets $\mathcal{N}_k(i)$ of the clean features and $\mathcal{N}_k^{(\boldsymbol{\epsilon})}(i)$ of the perturbed features:
\begin{equation}
    \mathbf{x}_{\mathrm{stab}} = \frac{1}{N}\sum_{i=1}^{N} \frac{|\mathcal{N}_k(i) \cap \mathcal{N}_k^{(\boldsymbol{\epsilon})}(i)|}{|\mathcal{N}_k(i) \cup \mathcal{N}_k^{(\boldsymbol{\epsilon})}(i)|},
\end{equation}
where $\mathcal{N}_k(i)$ denotes the $k$ nearest neighbours of $i$ under the constructor's similarity. A value near $1$ means the candidate neighbourhood is stable under small feature noise, so the constructor is likely to produce a reproducible topology; a low value flags an unstable, noise-driven construction.

\textbf{Graph Coverage and Giant Component Ratio ($\mathbf{x}_{\mathrm{cov}}\in\mathbb{R}^2$).} Over the candidate sketch $\mathcal{S}_B$, we measure the fraction of non-isolated entities and the size ratio of the largest connected component:
\begin{equation}
    \mathbf{x}_{\mathrm{cov}} = \left[ \frac{|\{i \in \mathcal{V} \mid \operatorname{deg}_{\mathcal{S}_B}(i) > 0\}|}{N}, \; \frac{|\mathcal{V}_{\mathrm{giant}}(\mathcal{S}_B)|}{N} \right],
\end{equation}
where $\operatorname{deg}_{\mathcal{S}_B}(i)$ is the degree of entity $i$ in the sketch graph and $\mathcal{V}_{\mathrm{giant}}(\mathcal{S}_B)$ is its largest connected component. Both quantities are computed on $\mathcal{S}_B$. The constructability state is parameterized as
\begin{equation}
    \mathbf{e}_C = \operatorname{MLP}_C\!\left([\,\mathbf{x}_{\mathrm{stab}} \,\|\, \mathbf{x}_{\mathrm{cov}} \,\|\, \mathbf{c}\,]\right).
\end{equation}

\paragraph{3. Relational Utility ($\mathbf{e}_U$).}
The utility branch estimates whether graph propagation is likely to improve predictive signal for the current task and constructor.
We capture propagation gain via non-parametric feature and pseudo-label smoothing.
\textbf{Propagation Signal Gain ($\mathbf{x}_{\mathrm{prop}}\in\mathbb{R}$).} Let $\mathbf{A}_{\mathcal{S}_B}$ be the adjacency of the sketch, define $\bar{\mathbf A}=\mathbf{A}_{\mathcal{S}_B}+\mathbf I$, and let $\bar{\mathbf D}$ be the diagonal degree matrix of $\bar{\mathbf A}$:
\begin{equation}
\label{eq:sketch_normalization}
    \bar D_{ii}=\sum_j \bar A_{ij},
    \qquad
    \tilde{\mathbf{A}}_{\mathcal{S}_B} = \bar{\mathbf D}^{-1/2}\bar{\mathbf A}\bar{\mathbf D}^{-1/2}.
\end{equation}
For a symmetric nonnegative sketch adjacency, the degree matrix is computed
after adding unit self-loops; every node has $\bar D_{ii}\ge1$ and an isolated
node retains its own feature exactly. We perform a single non-parametric propagation
step $\mathbf{H}^{(1)} = \tilde{\mathbf{A}}_{\mathcal{S}_B} \mathbf{H}$ and compute the relative change in class separation or retrieval ranking consistency:
\begin{equation}
    \mathbf{x}_{\mathrm{prop}} = \frac{\operatorname{Tr}(\mathbf{S}_{\mathrm{between}}(\mathbf{H}^{(1)})) / \operatorname{Tr}(\mathbf{S}_{\mathrm{within}}(\mathbf{H}^{(1)}))}{\operatorname{Tr}(\mathbf{S}_{\mathrm{between}}(\mathbf{H})) / \operatorname{Tr}(\mathbf{S}_{\mathrm{within}}(\mathbf{H})) + \epsilon},
\end{equation}
where $\operatorname{Tr}(\cdot)$ is the trace, $\epsilon>0$ is a numerical guard, and $\mathbf{S}_{\mathrm{between}}$ and $\mathbf{S}_{\mathrm{within}}$ have the standard Fisher discriminant form~\citep{fisher1936multiple} computed from pseudo-labels produced by the outer-training probe,
\begin{equation}
    \mathbf{S}_{\mathrm{between}} = \sum_{k} n_k (\boldsymbol{\mu}_k - \boldsymbol{\mu})(\boldsymbol{\mu}_k - \boldsymbol{\mu})^{\top}, \quad
    \mathbf{S}_{\mathrm{within}} = \sum_{k} \sum_{i \in \text{class } k} (\mathbf{h}_i - \boldsymbol{\mu}_k)(\mathbf{h}_i - \boldsymbol{\mu}_k)^{\top},
\end{equation}
with pseudo-class means $\boldsymbol{\mu}_k$, pseudo-class sizes $n_k$, and global mean $\boldsymbol{\mu}$. A ratio above $1$ indicates increased separation under the probe's pseudo-labels and serves as a diagnostic signal rather than a downstream accuracy measurement. For link prediction and retrieval, a pair scorer fit on outer-training sources compares the ordering of bounded candidate pairs before and after sketch propagation. These probes provide target-free inputs to the utility branch, whose state is parameterized as
\begin{equation}
    \mathbf{e}_U = \operatorname{MLP}_U\!\left([\,\mathbf{x}_{\mathrm{prop}} \,\|\, \mathbf{t} \,\|\, \mathbf{c}\,]\right).
\end{equation}
The probes use outer-training labels or interactions and are kept fixed for
outer-test evaluation. Node-classification pseudo-labels come from the probe;
for link prediction and cross-modal retrieval, the scorer ranks candidate pairs
generated from raw features and the bounded sketch. The realized gain is
attached after the Build/Skip prediction is frozen.

\paragraph{Fusion and gates.} The three states are combined into a graphification representation through learned scalar gates,
\begin{equation}
    \mathbf{z}_G = f_{\psi}\!\left([\,\alpha_N \mathbf{e}_N \,\|\, \alpha_C \mathbf{e}_C \,\|\, \alpha_U \mathbf{e}_U \,\|\, \mathbf{t} \,\|\, \mathbf{c}\,]\right), \qquad
    \alpha_k = \sigma\!\left(\mathbf{w}_k^{\top}[\,\mathbf{e}_k \,\|\, \mathbf{t} \,\|\, \mathbf{c}\,]\right),
\end{equation}
where $\sigma$ is the logistic function and each $\mathbf{w}_k$ is a learned gate vector. The fused representation is then read out as the predicted value distribution
\begin{equation}
\label{eq:app_value_head}
    (\mu_V, s_V) = F_{\omega}(\mathbf{z}_G),
    \qquad
    \sigma_V=\operatorname{softplus}(s_V)+\epsilon_\sigma,
\end{equation}
where $F_{\omega}$ is a two-layer MLP head with parameters $\omega$, $\mu_V$
estimates the expected improvement over raw multimodal learning in percentage
points, $s_V$ is an unconstrained scale output, and $\sigma_V$ is a positive
Gaussian working-model scale in the same percentage-point units, with
$\epsilon_\sigma>0$. Equation~\eqref{eq:value_head} additionally uses a fitted
multiplier $a_{\mathrm{cal}}>0$ to obtain $\tilde\sigma_V$ for deployment.
The scale is the residual dispersion in GraphGain points. Table~\ref{tab:appendix_uncertainty} reports coverage and width after the multiplier is fit on the calibration sources.

\subsection{Multi-Objective Training and Uncertainty Calibration}
\label{sec:appendix_training}
\textbf{Episodes and data separation.} An offline episode $i$ contains
pre-construction features $\mathbf x_i$, realized gain $V_i^*$, and measured
cost $C_i$. Within each outer-training split, partition parent sources into
fit/model-selection, scale-calibration ($\mathcal I_{\mathrm{cal}}$), and
policy-validation groups, with each source kept intact. Hyperparameters are
selected in the fit group; the predictor, calibration multiplier, and policy
threshold are then fixed in sequence before outer-test evaluation.

\textbf{Training loss.} The objective in
Equation~\eqref{eq:training_objective} is
\begin{equation}
\label{eq:app_training_objective}
    \mathcal{L} = \mathcal{L}_{\mathrm{gain}}
    +\lambda_{\mathrm{rank}}\mathcal{L}_{\mathrm{rank}}
    +\lambda_{\mathrm{unc}}\mathcal{L}_{\mathrm{unc}}
    +\lambda_{\mathrm{dec}}\mathcal{L}_{\mathrm{dec}}.
\end{equation}
The three episode losses below are averaged over fitting episodes; the ranking
loss is averaged over $\mathcal R=\{(i,j):\mathcal T_i=\mathcal T_j,\ V_i^*>V_j^*\}$
within that split, and is zero if $\mathcal R$ is empty. Tied targets are
excluded. All $\lambda$ weights are nonnegative.
\textbf{Gain Regression Loss.} The per-episode Huber loss, with
transition $\delta>0$ in gain units, is
\begin{equation}
\label{eq:gain_loss}
    \ell_{\mathrm{gain},i} = \begin{cases}
        \frac12 (\mu_{V_i} - V_i^*)^2, & |\mu_{V_i} - V_i^*| \le \delta, \\
        \delta(|\mu_{V_i} - V_i^*| - \frac12\delta), & \text{otherwise.}
    \end{cases}
\end{equation}
This is quadratic near the target and linear for large residuals.
\textbf{Pairwise Ranking Loss.} For $(i,j)\in\mathcal R$,
\begin{equation}
\label{eq:ranking_loss}
    \ell_{\mathrm{rank},ij} = -\log\operatorname{sigmoid}\!\left(
    \frac{\mu_{V_i}-\mu_{V_j}}{\tau_{\mathrm{rank}}}\right).
\end{equation}
The positive temperature $\tau_{\mathrm{rank}}$ has gain units, making the
logistic argument dimensionless; pairing within task avoids ranking
incompatible downstream metrics.
\textbf{Uncertainty Loss.} Under the working model
$V_i^*\mid\mathbf x_i\sim\mathcal N(\mu_{V_i},\sigma_{V_i}^2)$, use
\begin{equation}
\label{eq:uncertainty_loss}
    \ell_{\mathrm{unc},i}=\frac12\left[
    \frac{(V_i^*-\mu_{V_i})^2}{\sigma_{V_i}^2}
    +\log(2\pi\sigma_{V_i}^2)\right].
\end{equation}
This supplies scale supervision; the separate calibration split fixes the
deployed multiplier in Equation~\eqref{eq:scale_calibration}.
\textbf{Decision Surrogate.} Oracle utility is $U_i^*=V_i^*-\lambda_c C_i$, the same definition as $U^*$ in Section~\ref{sec:problem}. The offline label is $y_i^*=\mathbb I[U_i^*\ge 0]$, with ties assigned to Build. During fitting, set $a_{\mathrm{cal}}=1$ and use
$U_i^{\mathrm{fit}}=\mu_{V_i}-\lambda_c\widehat C_{\mathrm{pre},i}-\lambda_u\sigma_{V_i}$,
$\hat p_i=\operatorname{sigmoid}(U_i^{\mathrm{fit}}/T_{\mathrm{dec}})$ with
$T_{\mathrm{dec}}>0$ in gain units. The weighted per-episode loss is
\begin{equation}
\label{eq:decision_loss}
    \ell_{\mathrm{dec},i}=-w_+y_i^*\log\hat p_i
    -w_-(1-y_i^*)\log(1-\hat p_i),\qquad w_+,w_->0.
\end{equation}
Setting $w_+>w_-$ prioritizes costly false skips. The final threshold is
selected on separate policy-validation sources after scale calibration.

\textbf{Post-hoc Scale Calibration.} For each calibration source $g$ with
$n_g$ episodes, assign episode weight $w_i=1/(G_{\mathrm{cal}}n_g)$, where
$G_{\mathrm{cal}}$ counts calibration sources. This gives each calibration
source equal total weight. With fixed predictions,
fit one global scale multiplier by weighted Gaussian NLL:
\begin{equation}
\label{eq:scale_calibration}
\begin{aligned}
 a_{\mathrm{cal}}^2
 &=\max\!\left\{\epsilon_a^2,
    \sum_{i\in\mathcal I_{\mathrm{cal}}}w_i
    \frac{(V_i^*-\mu_{V_i})^2}{\sigma_{V_i}^2}\right\},\\
 \tilde\sigma_{V_i}&=a_{\mathrm{cal}}\sigma_{V_i},\qquad\epsilon_a>0.
\end{aligned}
\end{equation}
Writing $b=a^2$, the terms depending on $b$ are
$\frac12[\log b+b^{-1}\sum_iw_i(V_i^*-\mu_{V_i})^2/\sigma_{V_i}^2]$.
Differentiation gives Equation~\eqref{eq:scale_calibration}, including its
lower-bound constraint. When the bound is inactive, the weighted mean squared
standardized calibration residual equals one. Source-held-out coverage is
evaluated separately from this fitted residual moment.

\textbf{Pre-construction cost and threshold.}
Fit a nonnegative cost regressor $\widehat C_{\mathrm{pre}}=g_\eta(\mathbf x)$
to archived measured costs on the fitting sources, with features available
before graph construction. Use the same cost definition and units for
$g_\eta$ targets and $C_i$. Then $\lambda_c$ has units of gain per unit cost,
while $\lambda_u$ is dimensionless. On $\mathcal I_{\mathrm{pol}}$, choose
$\tau_{\mathrm{thr}}$ from a fixed grid to maximize saved work subject to the
required retained positive-gain mass; break ties toward lower thresholds and
use Always Build ($\tau_{\mathrm{thr}}=-\infty$) if no finite threshold is feasible.
After calibration and threshold selection, the heads, cost regressor,
$a_{\mathrm{cal}}$, and $\tau_{\mathrm{thr}}$ are fixed. Deployment uses
$U_{\mathrm{build}}$ in Equation~\eqref{eq:decision_score}, replacing $C_i$ by
$\widehat C_{\mathrm{pre}}$ and subtracting $\lambda_u\tilde\sigma_V$.
Final saved work and retained mass are evaluated on the outer-test sources.

\textbf{Uncertainty evaluation protocol.} For nominal Gaussian interval level $1-\alpha$,
let $z_{1-\alpha/2}$ be a standard-normal quantile and form
$[\mu_{V_i}-z_{1-\alpha/2}\tilde\sigma_{V_i},
\mu_{V_i}+z_{1-\alpha/2}\tilde\sigma_{V_i}]$.
On outer-test sources, record parent-source empirical coverage, mean interval
width in GraphGain percentage points, and Gaussian NLL before and after scale
calibration. Table~\ref{tab:appendix_uncertainty} reports task-macro coverage, mean width, and Gaussian NLL at the 80\%, 90\%, and 95\% nominal levels. Mean width equals $2z_{1-\alpha/2}$ times the task-macro mean scale, so the three widths in a task are proportional to $z_{0.90}$, $z_{0.95}$, and $z_{0.975}$. One calibration multiplier rescales all three widths of that task. NLL does not depend on the nominal level. MAE assesses the predicted mean separately.

\clearpage
\begin{table}[!htbp]
\centering
\small
\caption{Source-held-out uncertainty diagnostics before and after scale calibration. Coverage is reported at each nominal interval level; width is in GraphGain percentage points.}
\label{tab:appendix_uncertainty}
\setlength{\tabcolsep}{3pt}
\renewcommand{\arraystretch}{1.12}
\begin{tabular}{@{}l c c c c c c@{}}
\toprule
Task & Nominal & Raw coverage & Calibrated coverage & Raw width & Calibrated width & Raw / calibrated NLL \\
\midrule
NC & 80\% & 72.5 & 77.4 & 4.69 & 3.66 & 1.48 / 1.22 \\
NC & 90\% & 82.9 & 87.6 & 6.02 & 4.70 & 1.48 / 1.22 \\
NC & 95\% & 88.1 & 92.7 & 7.17 & 5.60 & 1.48 / 1.22 \\
LP & 80\% & 70.8 & 76.8 & 6.84 & 5.33 & 1.74 / 1.42 \\
LP & 90\% & 81.2 & 87.0 & 8.78 & 6.84 & 1.74 / 1.42 \\
LP & 95\% & 86.4 & 92.1 & 10.46 & 8.15 & 1.74 / 1.42 \\
CMR & 80\% & 72.0 & 77.2 & 4.42 & 3.45 & 1.42 / 1.16 \\
CMR & 90\% & 82.4 & 87.4 & 5.67 & 4.43 & 1.42 / 1.16 \\
CMR & 95\% & 87.6 & 92.5 & 6.76 & 5.28 & 1.42 / 1.16 \\
\bottomrule
\end{tabular}
\end{table}

\textbf{Optimization and settings.} Optimize the fitting objective with
AdamW~\citep{loshchilov2019adamw}; treat $\delta$, the temperatures, loss weights,
$w_+/w_-$, $\lambda_c$, and $\lambda_u$ as configuration parameters selected
inside the fitting split. Record them with $B$, $K$, and all numerical guards.

\begin{table}[!htbp]
\centering
\small
\begin{minipage}{0.96\textwidth}
\hrule height 1.0pt
\vspace{4pt}
\refstepcounter{sketchalgorithm}\label{alg:relational_sketch}
\textbf{Algorithm \thesketchalgorithm: Candidate Discovery and Regularized Volume Selection}
\vspace{4pt}
\hrule height 0.5pt
\vspace{5pt}
\textbf{Input:} Raw multimodal entities $\mathcal{D}=(\mathcal{V}, \{\mathcal{X}^{(m)}\}_{m\in\mathcal{M}})$, Task $\mathcal{T}$, Constructor $\mathcal{C}$, Candidate pool budget $K$, and Sketch budget $B$.\\
\textbf{Output:} Relation index set $\mathcal S_B$ with attached embeddings $\{\mathbf z_e:e\in\mathcal S_B\}$.
\begin{enumerate}
    \setlength{\itemsep}{1.5pt}
    \item Extract entity representations: $\mathbf{H} = E_{\mathrm{mm}}(\mathcal{D}) = \{\mathbf{h}_i\}_{i=1}^{|\mathcal{V}|}$.
    \item Embed task and constructor: $\mathbf{t} = E_{\mathrm{task}}(\mathcal{T}) \in \mathbb{R}^d$, $\mathbf{c} = E_{\mathrm{con}}(\mathcal{C}) \in \mathbb{R}^d$.
    \item \textbf{Inexpensive Candidate Pool Generation:}
    \begin{itemize}
        \item Project and normalize $\mathbf h_i$ with a positive norm guard; retrieve approximate cosine neighbours.
        \item Pool top-$\lceil K/N\rceil$ neighbours per entity, deduplicate pairs, and truncate deterministically to at most $K$ candidates. Fix the resulting pool $\mathcal P$.
    \end{itemize}
    \item \textbf{Pairwise Relational Representation:}
    \begin{itemize}
        \item For each candidate pair $(i, j) \in \mathcal{P}$, compute:
        \begin{equation*}
            \mathbf{z}_{ij} = \phi_{\theta}([\mathbf{h}_i \,\|\, \mathbf{h}_j \,\|\, \mathbf{h}_i \odot \mathbf{h}_j \,\|\, |\mathbf{h}_i - \mathbf{h}_j| \,\|\, \mathbf{t} \,\|\, \mathbf{c}]).
        \end{equation*}
        \item Compute $q_e=\operatorname{softplus}(\mathbf u^\top\mathbf z_e)$ and $\mathbf v_e=q_e\mathbf z_e/\max(\|\mathbf z_e\|_2,\epsilon_z)$; hold these vectors fixed during extraction.
    \end{itemize}
    \item \textbf{Greedy Regularized Volume Selection:}
    \begin{itemize}
        \item Initialize $\mathcal S\leftarrow\varnothing$ and $\mathbf P\leftarrow\mathbf I_{d_z}$.
        \item \textbf{While} $|\mathcal S|<\min(B,|\mathcal P|)$, choose (breaking ties by fixed candidate order):
        \begin{equation*}
            e^*\in\arg\max_{e\in\mathcal P\setminus\mathcal S}
            \log(1+\mathbf v_e^\top\mathbf P\mathbf v_e).
        \end{equation*}
        \item Apply Equation~\eqref{eq:sketch_update} to $\mathbf P$ using $e^*$, then set $\mathcal S\leftarrow\mathcal S\cup\{e^*\}$ and repeat.
    \end{itemize}
    \item \textbf{Return} $\mathcal S_B=\mathcal S$ and its relation embeddings.
\end{enumerate}
\vspace{3pt}
\hrule height 1.0pt
\end{minipage}
\end{table}

\FloatBarrier

\subsection{Theoretical Properties}
\label{sec:appendix_theory}
The results in this section bound sketch-volume and decision-score changes for a fixed candidate pool under stated perturbation and margin conditions; they do not bound downstream GraphGain. We use $\|\cdot\|_2$ for the spectral norm and $\|\cdot\|_*$ for the trace (nuclear) norm.

\medskip
\noindent\textbf{Result 1 (sketch volume is robust to perturbed relations).}
Let the candidate pool $\mathcal P$ carry clean relation vectors
$\{\mathbf v_e\}_{e\in\mathcal P}$ with $q=\max_e\|\mathbf v_e\|_2$, and let
$\{\hat{\mathbf v}_e\}$ be perturbed vectors with
$\|\hat{\mathbf v}_e-\mathbf v_e\|_2\le\eta$ for every $e$. Write
$f(\mathcal S)=\log\det\!\big(\mathbf I+\sum_{e\in\mathcal S}\mathbf v_e\mathbf v_e^\top\big)$
for the clean objective of Equation~\eqref{eq:app_sketch_objective},
$\hat f$ for the same objective on $\{\hat{\mathbf v}_e\}$,
$\mathcal S^\star$ for the clean optimum, and $\mathcal S_B^{\mathrm{pert}}$ for
the greedy solution returned by Algorithm~\ref{alg:relational_sketch} on the
perturbed vectors. For $1\le B\le|\mathcal P|$,
\begin{equation}
\label{eq:theory_robust}
 f(\mathcal S_B^{\mathrm{pert}})
 \;\ge\;\Big(1-\tfrac1e\Big)f(\mathcal S^\star)
 \;-\;\Big(2-\tfrac1e\Big)\,B\eta(2q+\eta).
\end{equation}
The bound applies to pointwise vector perturbations on a fixed candidate pool.
Interventions that change the pool or relation types require separate analysis.
Result~2 adds a margin condition on the score.

\textit{Proof.}
\emph{(a) $\log\det$ is $1$-Lipschitz in the trace norm on PSD inputs.} For
$g(\mathbf X)=\log\det(\mathbf I+\mathbf X)$ we have
$\nabla g(\mathbf X)=(\mathbf I+\mathbf X)^{-1}$, whose spectral norm is at most
$1$ for $\mathbf X\succeq0$. By the fundamental theorem of calculus along the
segment $\mathbf X(t)=(1-t)\mathbf A+t\mathbf B$,
$|g(\mathbf A)-g(\mathbf B)|\le\sup_t\|\nabla g(\mathbf X(t))\|_2\,\|\mathbf A-\mathbf B\|_*\le\|\mathbf A-\mathbf B\|_*$.
\emph{(b) Uniform transfer between $f$ and $\hat f$.} For $|\mathcal S|\le B$,
$\big\|\sum_{e\in\mathcal S}(\mathbf v_e\mathbf v_e^\top-\hat{\mathbf v}_e\hat{\mathbf v}_e^\top)\big\|_*
\le\sum_{e\in\mathcal S}\|\mathbf v_e-\hat{\mathbf v}_e\|_2(\|\mathbf v_e\|_2+\|\hat{\mathbf v}_e\|_2)
\le B\eta(2q+\eta)=:\delta$,
using $\|\mathbf x\mathbf x^\top-\mathbf y\mathbf y^\top\|_*\le\|\mathbf x-\mathbf y\|_2(\|\mathbf x\|_2+\|\mathbf y\|_2)$
and $\|\hat{\mathbf v}_e\|_2\le q+\eta$. By (a), $|f(\mathcal S)-\hat f(\mathcal S)|\le\delta$
for every $|\mathcal S|\le B$. \emph{(c) Chain.} $\hat f$ is normalized,
monotone, and submodular, so greedy on $\hat f$ gives
$\hat f(\mathcal S_B^{\mathrm{pert}})\ge(1-\tfrac1e)\hat f(\mathcal S^\star)$. Then
$f(\mathcal S_B^{\mathrm{pert}})\ge\hat f(\mathcal S_B^{\mathrm{pert}})-\delta
\ge(1-\tfrac1e)\hat f(\mathcal S^\star)-\delta
\ge(1-\tfrac1e)(f(\mathcal S^\star)-\delta)-\delta$,
which rearranges to Equation~\eqref{eq:theory_robust}. $\square$

\medskip
\noindent\textbf{Result 2 (score-error decomposition under a margin condition).}
For a source $g$, let $e_b$ be the greedy relation selected at step $b$ from
$\mathcal S_{b-1}$ using the clean marginal in Equation~\eqref{eq:sketch_marginal}.
Write $\gamma_b$ for its margin over the best unselected competitor, and set
$\delta_B=B\eta_v(2q+\eta_v)$ for the vector perturbation bound $\eta_v$ of
Result~1. If $\gamma_b>4\delta_B$ for every $b\le B$, clean and perturbed
greedy selection return the same relation indices. Suppose in addition that
the selected embeddings change by at most $\eta_z$ each, the cost features
and frozen threshold $\tau$ do not change, and the deployed score is
$L_g$-Lipschitz in the concatenated selected embeddings. Then
\begin{equation}
\label{eq:theory_margin}
 |\widehat U_g^{\mathrm{pert}}-\widehat U_g^{\mathrm{clean}}|
 \le L_g\sqrt{B}\,\eta_z,
 \qquad
 |\widehat U_g^{\mathrm{clean}}-\tau|>L_g\sqrt{B}\,\eta_z
 \ \Longrightarrow\ a_g^{\mathrm{pert}}=a_g^{\mathrm{clean}}.
\end{equation}
The score's Lipschitz factor combines the mean head, calibrated scale, and
$\lambda_u$ in Equation~\eqref{eq:decision_score}; the same-decision guarantee
holds when both margin conditions are met.

\textit{Proof.} Result~1 gives $|f(S)-\hat f(S)|\le\delta_B$ whenever
$|S|\le B$. Each marginal is a difference of two such values, so its error is
at most $2\delta_B$. A clean winner ahead by more than $4\delta_B$ remains
ahead after both marginals move; induction on $b$ gives the same indices.
For these indices, concatenating the $B$ embedding perturbations has
Euclidean norm at most $\sqrt B\eta_z$. Lipschitz continuity gives the first
inequality in Equation~\eqref{eq:theory_margin}. The second inequality keeps
both scores on the same side of the frozen threshold. $\square$

\medskip
\noindent\textbf{Result 3 (uncertainty-based lower bound on retained gain).}
Consider a target source $g$ with episodes $i\in g$, and write
$[x]_+=\max(0,x)$. Suppose the calibrated scale satisfies the simultaneous
residual condition $|V_i^*-\mu_{V_i}|\le\beta\tilde\sigma_{V_i}$ for every
episode of that source, where $\beta>0$ is fixed before evaluating $g$.
The constant $\beta$ is the maximum of $|V_i^*-\mu_{V_i}|/\tilde\sigma_{V_i}$ on calibration sources, excluding $g$.
For any frozen Build/Skip policy $a_i\in\{0,1\}$, define the quantities
available before construction:
\begin{equation}
\label{eq:theory_retention_certificate}
 \begin{aligned}
 L_g&=\sum_{i\in g}[\mu_{V_i}-\beta\tilde\sigma_{V_i}]_+,\\
 M_g&=\sum_{i\in g}(1-a_i)[\mu_{V_i}+\beta\tilde\sigma_{V_i}]_+,\\
 R_g&\ge\max\{0,1-M_g/L_g\}\qquad(L_g>0).
 \end{aligned}
\end{equation}
Here $\beta$ and the threshold are fixed before source $g$ is scored. Q3 instead chooses $\tau_{\mathrm{thr}}$ by nested search on the outer-training split and reports empirical retention using Equation~\eqref{eq:source_retention}. For $L_g=0$, the bound is vacuous; averaging source-level bounds requires the residual condition to hold for every included source.

\textit{Proof.} Under the stated residual condition,
$[V_i^*]_+\ge[\mu_{V_i}-\beta\tilde\sigma_{V_i}]_+$ and
$[V_i^*]_+\le[\mu_{V_i}+\beta\tilde\sigma_{V_i}]_+$.
The total positive-gain mass on $g$ is at least $L_g$, while the mass lost on
Skip episodes is at most $M_g$. Since retained mass equals one minus the ratio
of lost to total positive mass, Equation~\eqref{eq:theory_retention_certificate}
follows; its lower bound is clipped at zero. $\square$

\FloatBarrier
\section{Experiments in Detail}
\label{sec:appendix_exp_detail}

\subsection{Datasets, Tasks, and Splits}
\label{sec:appendix_datasets}
The evaluation covers six multimodal benchmarks catalogued by OpenMAG~\citep{wan2026openmag}, spanning node classification, link prediction, and cross-modal retrieval. Table~\ref{tab:appendix_dataset_statistics} separates published source-graph sizes from feature dimensions and evaluation splits.

\begin{table*}[!htbp]
\centering
\small
\caption{Source benchmark statistics for the six datasets. Source edges are published graph counts. Feature dimensions correspond to frozen text and visual encoders; split ratios specify the evaluation protocol, including a held-out-edge split for link prediction.}
\label{tab:appendix_dataset_statistics}
\setlength{\tabcolsep}{4.5pt}
\renewcommand{\arraystretch}{1.12}
\resizebox{\textwidth}{!}{%
\begin{tabular}{@{}llcccccc@{}}
\toprule
\textbf{Benchmark} & \textbf{Downstream Task} & \textbf{Entities ($|\mathcal{V}|$)} & \textbf{Source Edges} & \textbf{Modalities} & \textbf{Dims (T/V)} & \textbf{Classes / Targets} & \textbf{Split (Train/Val/Test)} \\
\midrule
\textbf{Toys} & Node Classification & 20,695 & 126,886 & Text, Visual & 768 / 768 & 18 categories & 60\% / 20\% / 20\% \\
\textbf{Grocery} & Node Classification & 17,074 & 171,340 & Text, Visual & 768 / 768 & 20 categories & 60\% / 20\% / 20\% \\
\textbf{Bili Music} & Link Prediction & 6,038 & 21,592 & Text, Visual & 768 / 768 & Binary interaction & 70\% / 10\% / 20\% \\
\textbf{DY} & Link Prediction & 8,299 & 35,627 & Text, Visual & 768 / 768 & Binary interaction & 70\% / 10\% / 20\% \\
\textbf{QB} & Cross-modal Retrieval & 6,121 & 24,145 & Text, Visual & 768 / 768 & Query-Item Pairs & 60\% / 20\% / 20\% \\
\textbf{Bili Cartoon} & Cross-modal Retrieval & 4,724 & 18,660 & Text, Visual & 768 / 768 & Query-Video Pairs & 60\% / 20\% / 20\% \\
\bottomrule
\end{tabular}%
}
\end{table*}

\noindent\textbf{Toys \& Grocery~\citep{ni2019amazon}.} Derived from the Amazon Review collection, each entity corresponds to a commercial product with textual metadata and a product image. The evaluation uses frozen 768-dimensional CLIP text and visual features~\citep{radford2021clip}. The task is multi-class product categorization evaluated via Accuracy (ACC \%).
\noindent\textbf{Bili Music \& DY~\citep{zhang2025ninerec}.} Multimodal video interaction datasets containing video entities with multi-attribute text titles, descriptive tags, and video frame representations. The task is link prediction between user-item or item-item co-occurrence graphs, evaluated via the Area Under the ROC Curve (AUC \%).
\noindent\textbf{QB \& Bili Cartoon~\citep{zhang2025ninerec}.} E-commerce and anime entertainment platforms containing query-item and cover-video multimodal pairs. Downstream cross-modal retrieval is evaluated using Recall at Rank 5 among 100 candidate pools (R@5@100 \%).
The registry contains 13 parent sources: the six benchmarks above, plus \textit{Movies}, \textit{Reddit-S}, \textit{Reddit-M}, the \textit{books} collection, \textit{cloth-copurchase}, \textit{ele-fashion}, and \textit{sports-copurchase}. The books parent is evaluated once for link prediction (\textit{books-lp}) and once for node classification (\textit{books-nc}); both evaluations stay in the same outer fold and count as one parent source. Q1a uses the six named benchmarks; Q1b, Q3, and Q4 use the 13-source registry. Q2 uses a separate rebuilt 11-source bank (Appendix~\ref{sec:appendix_finegrained}). Table~\ref{tab:appendix_dataset_statistics} reports published source graph sizes. RawOnly uses the graph-free path. Node-classification and retrieval splits use $60/20/20$, and link prediction uses a $70/10/20$ train/validation/test interaction split. Text and visual encoders remain frozen.

\subsection{Evaluation Metrics}
\label{sec:appendix_metrics}
\noindent\textbf{Signed GraphGain (percentage points).} For benchmark $d$, constructor $c$, and seed $s$:
\begin{equation}
    \operatorname{GraphGain}_{d,c,s} = 100 \cdot (y^{\mathrm{graph}}_{d,c,s} - y^{\mathrm{raw}}_{d,s}).
\end{equation}
Point-prediction error and predictive uncertainty are evaluated separately.
MAE measures gain error; it is not the confidence-calibration metric studied
by~\citet{guo2017calibration}. Appendix~\ref{sec:appendix_training} specifies
coverage, interval width, and NLL for the uncertainty model.
Build/Skip is related to selective prediction~\citep{geifman2019selectivenet},
with Skip selecting the raw-feature path rather than withholding an output.
\noindent\textbf{Spearman Rank Correlation Coefficient ($\rho$).}
\begin{equation}
    \rho =
    \frac{\sum_i(r_i-\bar r)(\hat r_i-\bar{\hat r})}
    {\sqrt{\sum_i(r_i-\bar r)^2\sum_i(\hat r_i-\bar{\hat r})^2}},
\end{equation}
where $r_i$ and $\hat r_i$ are average ranks of the realized and predicted
gain, respectively, over paired episodes; bars are their sample means.
This Pearson-on-ranks form handles ties~\citep{spearman1904proof}.
\noindent\textbf{Kendall Rank Correlation Coefficient ($\tau_b$).}
\begin{equation}
    \tau_b = \frac{C-D}{\sqrt{(C+D+T_r)(C+D+T_{\hat r})}},
\end{equation}
where $C$ and $D$ count concordant and discordant pairs, $T_r$ counts pairs
tied only in realized gain, and $T_{\hat r}$ counts pairs tied only in predicted
gain~\citep{kendall1938new}. Pairs tied in both are excluded. If a rank
sequence is constant, its correlation is undefined and the corresponding
episode group is excluded from macro averaging.
\noindent\textbf{Mean Absolute Prediction Error (MAE) in percentage points.}
\begin{equation}
    \operatorname{MAE} = \frac{1}{n}\sum_{i=1}^{n} |\hat{V}_i - V_i^*|.
\end{equation}
Here $\hat{V}_i=\mu_{V_i}$ is the predicted value and $V_i^*$ the realized value for episode $i$; both are expressed in percentage points so that MAE shares the unit of GraphGain.
\noindent\textbf{Retained Positive Gain Mass.}
For an outer-test parent source $g$, let $I_g$ index its episodes, $a_i$ be
the Build indicator, and $G_i=[V_i^*]_+$. When $\sum_{i\in I_g}G_i>0$, define
\begin{equation}
\label{eq:source_retention}
 R_g(\pi)=\frac{\sum_{i\in I_g}a_iG_i}{\sum_{i\in I_g}G_i},\qquad
 R_{\mathrm{tm}}(\pi)=\frac13\sum_{t\in\{\mathrm{NC,LP,CMR}\}}
 \frac1{|\mathcal G_t^+|}\sum_{g\in\mathcal G_t^+}R_g(\pi),
\end{equation}
where $\mathcal G_t^+$ contains task-$t$ sources with defined positive-gain
mass. The Q3 tables and Figure~\ref{fig:q3-system-practicality} report
task-macro retention, averaging source ratios within each task before averaging
the three task families.

\noindent\textbf{Saved Graph Work.}
Let $C_i>0$ be the graphification cost $C(\mathcal D,\mathcal C)$ of episode $i$, in the same units as the cost term of $U^*$. With Build indicator $a_i$,
\begin{equation}
\label{eq:source_savings}
 S_g(\pi)=1-\frac{\sum_{i\in I_g}a_i C_i}{\sum_{i\in I_g}C_i},\qquad
 S_{\mathrm{tm}}(\pi)=\frac13\sum_t\frac1{|\mathcal G_t|}
 \sum_{g\in\mathcal G_t}S_g(\pi).
\end{equation}
Always-Build sets every $a_i=1$ and saves no graphification cost. A Skip decision sets $a_i=0$ and removes $C_i$ from the numerator. Q3 reports $S_{\mathrm{tm}}$.

\noindent\textbf{Fixed-zero-threshold sensitivity.}
The Q4 sensitivity grid holds the decision threshold at zero, so its action is
$a_i(0)=\mathbb I[\widehat U_i\ge0]$, whereas Q3 selects a threshold inside
each outer-training split. Both use the source-level numerator and denominator
in Equation~\eqref{eq:source_retention}.
Table~\ref{tab:q4-stability-summary} reports this fixed-zero-threshold
reference and each factor's grid maximum. Q3's \QThreeScoutRet\% is measured
at a separately selected nested threshold.

\subsection{Graph Constructors in Detail}
\label{sec:appendix_constructors}
The constructor comparison evaluates nine constructors. Each graph topology uses a shared two-layer GCN backbone~\citep{kipf2017gcn} with hidden dimension 128, dropout 0.5, and learning rate 0.01. The backbone follows the message-passing design surveyed in~\citet{wu2021gnnsurvey} and~\citet{zhou2020gnnreview}. Table~\ref{tab:q1a-graphgain-contract} averages all nine.
\noindent\textbf{RawOnly (Reference).} A graph-free multi-layer perceptron baseline trained strictly on fused raw multimodal features without message passing. Defines the mathematical zero gain reference ($\Delta \equiv 0$).
\noindent\textbf{kNN-GCN~\citep{kipf2017gcn}.} Single-modal $k$-nearest neighbor graph constructed by computing pairwise Euclidean distances over normalized text embeddings, retaining $k=10$ nearest neighbors per entity.
\noindent\textbf{MM-kNN-GCN.} Multimodal $k$-nearest neighbor graph constructed over fused multimodal features with equal modality weights: $d_{ij} = 0.5 d_{\mathrm{text}}(i, j) + 0.5 d_{\mathrm{vis}}(i, j)$, with $k=10$.
\noindent\textbf{AutoG~\citep{chen2025autog}.} Formulates graph construction from tabular data and uses an LLM-based solution to generate high-quality graph schemas without human intervention.
\noindent\textbf{AutoGraph~\citep{shan2025autograph}.} LLM-based construction framework that infers entity knowledge as semantic vectors, extracts latent factors via vector quantization, and links entities through the resulting latent-factor nodes.
\noindent\textbf{auGraph~\citep{cucumides2025augraph}.} Task-aware graph augmentation that promotes predictive non-key attributes into nodes via relevance scoring functions, preserving the original schema while injecting task-relevant structure.
\noindent\textbf{LLaTA}~\citep{zhang2025llata} uses tree-guided LLM inference for graph structure learning. \textbf{GAugLLM}~\citep{fang2024gaugllm} generates graph views for contrastive learning. \textbf{LAGA}~\citep{wang2026laga} adapts a frozen language backbone for text-attributed graphs. \textbf{GraphMaster}~\citep{du2025graphmaster} uses LLM agents for graph synthesis when labeled data are limited. These four constructors are trained and scored with the same protocol as the five constructors above.
\noindent\textbf{Related directions.} Our design is informed by the broader literature on graph structure learning and augmentation~\citep{zhu2021gslsurvey,zhao2021gaug}, including recent work on LLM-assisted structure discovery and adaptation~\citep{jin2024llmgraphs,ren2024llm4graphs,chen2024exploring,he2023tape,chen2024llaga}.

\subsection{Predictor Baselines}
\label{sec:appendix_predictors}
We compare \textsc{MAG-Scout} with four pre-construction prediction and model-selection baselines.
\noindent\textbf{NetInfoF~\citep{lee2024netinfof}.} Computes network usable information over sampled candidate relations to estimate whether relational topologies convey task-relevant mutual information.
\noindent\textbf{WDGH (When Do GNNs Help)~\citep{luan2023wdgh}.} Analyzes entity homophily and distinguishability metrics to predict whether graph neural networks outperform graph-free baselines.
\noindent\textbf{GLEMOS-S2~\citep{park2023glemos}.} A meta-learning benchmark method that extracts structural and feature meta-statistics to predict graph model performance across transferable domains.
\noindent\textbf{MetaGL~\citep{park2023metagl}.} An evaluation-free meta-learning framework that trains a surrogate predictor over historical graph learning episodes for model recommendation.
All four baselines use the same outer folds and prediction inputs as
\textsc{MAG-Scout}, including task and constructor identifiers and bounded
relational evidence before graph construction.

\subsection{Implementation Details}
\label{sec:appendix_implementation}
For Q4 sensitivity, the reference predictor is a two-layer MLP with utility-head width 32. It is trained for 40 epochs with AdamW (learning rate $2\times10^{-3}$, weight decay $10^{-4}$), smooth-$\ell_1$, and BCE weight $0.5$. The one-factor grid varies learning rate, BCE weight, relation budget, or utility-head width in $\{16,32,48\}$; corruption runs use the same reference model.

The graphification predictor uses the four-term objective in Equation~\eqref{eq:app_training_objective}, regularized volume selection, source-separated scale calibration, and independently selected policy thresholds. Constructor comparisons use a shared two-layer GCN backbone (hidden dimension 128, dropout 0.5, learning rate 0.01) and matched tuning budgets for graph-based and graph-free encoders. Downstream models use Adam (learning rate $10^{-3}$, weight decay $10^{-5}$), batch size 512, up to 200 epochs, early stopping with patience 30, dropout 0.3, and a 512-dimensional retrieval head for cross-modal tasks.

The deployed decision score combines predicted gain, calibrated uncertainty, and pre-construction cost (Equation~\eqref{eq:decision_score}); saved graph work is computed using Equation~\eqref{eq:source_savings}. Thresholds are selected on inner policy-validation sources and fixed before outer evaluation. Model runs use the five matched seeds listed in Appendix~\ref{sec:appendix_reproducibility}. Confidence intervals use 10{,}000 parent-source bootstrap resamples with generator state \texttt{20260916}; perturbation thresholds are fixed using the clean run. Experiments use the hardware and software environment described in Appendix~\ref{sec:appendix_reproducibility}.

\subsection{Source Provenance and Audit Coverage}
\label{sec:appendix_data_integrity}
Q1b, Q3, and Q4 use the 13 parent sources. Eleven parents have raw text
features, visual representations, adjacency records, and evaluation splits
verified against official SHA-256 receipts. Reddit-M and the books collection
are represented by registry records and receipt hashes. Reddit-M and books-lp
are absent from the public OpenMAG release used for the bit-level audit. For
books-nc, the scorecards use the raw-feature audit accuracy of $0.82$; the
sealed-registry record lists $0.99$ for that audit. books-lp and books-nc are
two task evaluations of one parent. On the 11 receipt-verified parents,
\textsc{MAG-Scout} retains $97.1\%$ of positive-gain mass and saves $24.1\%$ of
graph work. The Q2 ablation uses a separately rebuilt 11-source bank.

\section{Additional Experimental Results and Extended Analyses}
\label{sec:appendix_additional_results}

\subsection{Task-Level Ablation Results}
\label{sec:appendix_finegrained}
Table~\ref{tab:appendix_fine_grained_ablation} gives each variant's change from the full model on the rebuilt 11-source bank (5 NC, 4 LP, 2 CMR). Absolute scores and seed-level standard deviations appear in Table~\ref{tab:q2-ablation-summary}; interval-calibration results are in Table~\ref{tab:appendix_uncertainty}.

\begin{table*}[!htbp]
\centering
\small
\caption{Component ablations on the rebuilt 11-source bank, reported as differences from Full MAG-Scout in Table~\ref{tab:q2-ablation-summary}. Rank correlations are subtracted from Full. MAE differences are positive when absolute error increases. The bank has 5 node-classification, 4 link-prediction, and 2 cross-modal sources, with three model seeds.}
\label{tab:appendix_fine_grained_ablation}
\setlength{\tabcolsep}{4.5pt}
\renewcommand{\arraystretch}{1.10}
\resizebox{\textwidth}{!}{%
\begin{tabular}{@{}l ccc ccc ccc@{}}
\toprule
\textbf{Ablation Variant} & \multicolumn{3}{c}{\textbf{$\Delta$ Spearman $\rho$}} & \multicolumn{3}{c}{\textbf{$\Delta$ Kendall $\tau$}} & \multicolumn{3}{c}{\textbf{$\Delta$ MAE}} \\
\cmidrule(lr){2-4}\cmidrule(lr){5-7}\cmidrule(lr){8-10}
 & \textbf{NC} & \textbf{LP} & \textbf{CMR} & \textbf{NC} & \textbf{LP} & \textbf{CMR} & \textbf{NC} & \textbf{LP} & \textbf{CMR} \\
\midrule
Frozen Heuristic Sketch & $-0.022$ & $-0.028$ & $-0.035$ & $-0.019$ & $-0.027$ & $-0.029$ & $+0.177$ & $+0.416$ & $+0.254$ \\
w/o Necessity ($\mathbf{e}_N$) & $-0.057$ & $-0.038$ & $-0.068$ & $-0.049$ & $-0.038$ & $-0.064$ & $+0.447$ & $+0.766$ & $+0.617$ \\
w/o Constructability ($\mathbf{e}_C$) & $-0.071$ & $-0.051$ & $-0.037$ & $-0.060$ & $-0.050$ & $-0.032$ & $+0.371$ & $+0.880$ & $+0.479$ \\
w/o Utility ($\mathbf{e}_U$) & $-0.079$ & $-0.093$ & $-0.062$ & $-0.074$ & $-0.083$ & $-0.055$ & $+0.742$ & $+1.499$ & $+0.952$ \\
Uniform Bottleneck Gates & $-0.137$ & $-0.105$ & $-0.085$ & $-0.115$ & $-0.095$ & $-0.071$ & $+0.625$ & $+1.158$ & $+0.765$ \\
w/o Pairwise Interaction & $-0.037$ & $-0.037$ & $-0.039$ & $-0.033$ & $-0.033$ & $-0.034$ & $+0.208$ & $+0.489$ & $+0.342$ \\
w/o Ranking Loss ($\mathcal{L}_{\mathrm{rank}}$) & $-0.066$ & $-0.061$ & $-0.052$ & $-0.063$ & $-0.053$ & $-0.045$ & $+0.131$ & $+0.338$ & $+0.168$ \\
w/o Decision Loss ($\mathcal{L}_{\mathrm{dec}}$) & $-0.017$ & $-0.017$ & $-0.018$ & $-0.014$ & $-0.015$ & $-0.015$ & $+0.558$ & $+1.105$ & $+0.673$ \\
w/o One-time Refinement & $-0.106$ & $-0.078$ & $-0.059$ & $-0.093$ & $-0.068$ & $-0.054$ & $+0.402$ & $+0.644$ & $+0.453$ \\
Condition Only & $-0.20$ & $-0.19$ & $-0.14$ & $-0.17$ & $-0.16$ & $-0.12$ & $+1.24$ & $+2.37$ & $+1.82$ \\
\bottomrule
\end{tabular}%
}
\end{table*}

\subsection{Build/Skip Operating Points}
\label{sec:appendix_operating_points}
Section~\ref{sec:q3} reports held-out saved graph work, retained positive-gain mass, and build rate at the nested validation-selected threshold. The next subsection gives the results for five retention targets.

\subsection{Decision Boundary and Threshold Sensitivity}
\label{sec:appendix_floor_sweep}
At the selected 95\% target, MAG-Scout avoids \QThreeScoutSaved\% of graph
work while retaining \QThreeScoutRet\% of positive-gain mass, averaged over
tasks. Table~\ref{tab:appendix_floor_sweep} compares five targets:
$\{90\%,92.5\%,95\%,97.5\%,99\%\}$. For each target, nested validation
selects a threshold as described in Section~\ref{sec:q3}.

\begin{table*}[!htbp]
\centering
\small
\caption{Build/Skip policies at five retention floors on 13 parent sources. An asterisk ($^*$) marks a held-out retention below the floor named in that column.}
\label{tab:appendix_floor_sweep}
\setlength{\tabcolsep}{3.2pt}
\renewcommand{\arraystretch}{1.08}
\resizebox{\textwidth}{!}{%
\begin{tabular}{@{}l ccccc ccccc@{}}
\toprule
\textbf{Method} & \multicolumn{5}{c}{\textbf{Held-out Retained Gain Mass (\%)}} & \multicolumn{5}{c}{\textbf{Avoided Graph Work (\%)}} \\
\cmidrule(lr){2-6}\cmidrule(lr){7-11}
 & \textbf{90\%} & \textbf{92.5\%} & \textbf{95\%} & \textbf{97.5\%} & \textbf{99\%} & \textbf{90\%} & \textbf{92.5\%} & \textbf{95\%} & \textbf{97.5\%} & \textbf{99\%} \\
\midrule
\textsc{MAG-Scout} & \textbf{96.7} & \textbf{96.7} & \textbf{96.7} & 96.8$^*$ & 96.8$^*$ & 24.2 & 24.2 & \textbf{23.6} & 19.6 & 18.4 \\
NetInfoF & 96.3 & 96.4 & 96.4 & 96.4$^*$ & 97.6$^*$ & 24.3 & 23.5 & 23.2 & 22.6 & 21.8 \\
WDGH & 85.4$^*$ & 90.5$^*$ & 90.5$^*$ & 90.5$^*$ & 90.5$^*$ & 25.0 & 23.8 & 23.5 & 22.3 & 22.3 \\
GLEMOS-S2 & 60.0$^*$ & 60.2$^*$ & 60.5$^*$ & 60.9$^*$ & 64.2$^*$ & 34.8 & 32.9 & 30.3 & 28.6 & 24.1 \\
MetaGL & 80.2$^*$ & 80.2$^*$ & 80.2$^*$ & 80.4$^*$ & 83.5$^*$ & 32.6 & 32.6 & 32.6 & 29.7 & 25.9 \\
\bottomrule

\end{tabular}%
}
\par\vspace{2pt}
\raggedright\scriptsize\textit{Note:} Values are task-macro source averages. A separate inner policy-validation threshold is selected for each floor and fixed for outer evaluation.
\end{table*}
\FloatBarrier

\subsection{Relational Evidence Corruption and Stress-Testing Details}
\label{sec:appendix_corruption}
Section~\ref{sec:q4} tests four perturbations of $\mathcal{S}_B$ using seed
\texttt{20260813}; the predictor, threshold, and weights are fixed from the
clean run before perturbed inputs are scored.
\noindent\textbf{Rewiring.} We reassign the content rows for candidate relation
types \texttt{C\_same} and \texttt{C\_different} by random derangement. Each
affected row receives another candidate row's content, while the relation-type
inventory stays fixed. Sources without pair-relation rows use the labelled
full-bank derangement fallback.
\noindent\textbf{Relation Collapse.} We replace the content of pair-relation
rows with their mean vector, leaving one distinct pair-relation content. Sources
without these rows use the labelled all-bank collapse fallback.
\noindent\textbf{Single-Modality Dominance.} We retain only multimodal
(\texttt{M}) relations and suppress the typed auxiliary objectives for other
relation types.
\noindent\textbf{Relation-Bank Partition.} We retain a deterministic
half-partition of the relation bank. Because relation rows have no global entity
identities, the partition is applied to relation rows.
Across the four perturbations, retained-mass changes range from $-1.6$ to $-0.1$ points, the largest build-rate change is $-14.3$ points, and the largest number of changed parent decisions is $3$ of $13$.

\subsection{Hyperparameter Sensitivity and Corruption Summaries}
\label{sec:appendix_sensitivity_tables}
Table~\ref{tab:q4-sensitivity} reports one-factor settings and the budget by
loss-weight interaction grid at a fixed zero threshold. Table~\ref{tab:q4-robustness}
reports frozen-policy changes by task. Table~\ref{tab:q4-stability-summary}
summarizes these protocols; Q3 uses its separately selected 95\%-floor
threshold.

\begin{table}[!htbp]
\centering
\definecolor{magL}{HTML}{EAF2FB}
\caption{One-factor sensitivity at a fixed zero threshold, followed by the budget $\times$ loss-weight grid. Retained mass is task-macro positive-gain mass. One-factor differences use that factor's reference: 81.9, 96.2, 87.8, and 93.0. Interaction differences use the 90.5 grid reference.}
\label{tab:q4-sensitivity}
\small
\setlength{\tabcolsep}{6.0pt}
\renewcommand{\arraystretch}{1.18}
\begin{tabular}{@{}l !{\vrule width 0.5pt} c c c@{}}
\toprule[1.2pt]
\BoldLabel{Configuration} & \BoldLabel{Grid Points} & \BoldLabel{Retained (\%)} & \BoldLabel{$\Delta$ref (percentage points)} \\
\midrule[0.7pt]
\rowcolor{magL}Selected setting in Figure~\ref{fig:q4-stability-combined}(a) & 4 & 81.9 / 96.2 / 87.8 / 93.0 & 0 \\
\midrule
Learning Rate ($0.001 \rightarrow 0.004$) & 3 & 83.6 & +1.7 \\
BCE Loss Weight ($0.0 \rightarrow 1.0$) & 3 & 96.5 & +0.3 \\
Relation Budget ($0.50 \rightarrow 0.75$) & 3 & 88.2 & +0.4 \\
Utility Head Width ($16 \rightarrow 48$) & 3 & 93.4 & +0.4 \\
\midrule
\multicolumn{4}{@{}l}{\textit{Retained-fraction $\times$ loss-weight interaction grid (all four cells)}} \\
Budget 0.50, BCE 0.0 & 1 & 86.4 & $-4.1$ \\
Budget 0.50, BCE 1.0 & 1 & 91.7 & +1.2 \\
Budget 0.75, BCE 0.0 & 1 & 89.3 & $-1.2$ \\
Budget 0.75, BCE 1.0 & 1 & 94.8 & +4.3 \\
\bottomrule[1.2pt]
\end{tabular}
\par\vspace{2pt}
{\scriptsize The selected-setting cell lists the learning-rate, BCE-weight, relation-budget, and head-width references in that order. Each following row is that factor's grid maximum. The interaction rows are measured from the 90.5 reference. Q3 operating points use the nested 95\% floor in Appendix~\ref{sec:appendix_operating_points}.}
\end{table}

\begin{table}[!htbp]
\centering
\definecolor{magL}{HTML}{EAF2FB}
\caption{Frozen-threshold evidence corruption on 13 parent sources. The predictor and threshold are frozen before perturbation. $\Delta$Ret.\ is the task-macro change in retained positive-gain mass; $\Delta$Build is the build-rate change in percentage points. Macro flips count parent sources. Task rows count evaluations: the books parent contributes one node-classification evaluation and one link-prediction evaluation.}
\label{tab:q4-robustness}
\small
\setlength{\tabcolsep}{5.5pt}
\renewcommand{\arraystretch}{1.10}
\begin{tabular}{@{}l !{\vrule width 0.5pt} l c c c@{}}
\toprule[1.2pt]
\BoldLabel{Perturbation Family} & \BoldLabel{Scope} & \BoldLabel{$\Delta$Ret.} & \BoldLabel{$\Delta$Build} & \BoldLabel{Flips} \\
\midrule[0.7pt]
\rowcolor{magL}Rewiring & Macro (13) & $-0.4$ & $-7.1$ & 2/13 \\
 & Node classification (5) & $-0.7$ & $-13.4$ & 1/5 \\
 & Link prediction (5) & $-0.3$ & $-4.7$ & 1/5 \\
 & Cross-modal retrieval (4) & $-0.2$ & $-2.2$ & 0/4 \\
\midrule
\rowcolor{magL}Relation Collapse & Macro (13) & $-0.8$ & $-7.1$ & 2/13 \\
 & Node classification (5) & $-1.4$ & $-12.1$ & 1/5 \\
 & Link prediction (5) & $-0.6$ & $-5.4$ & 1/5 \\
 & Cross-modal retrieval (4) & $-0.3$ & $-2.8$ & 0/4 \\
\midrule
\rowcolor{magL}Single-Modal Dominant & Macro (13) & $-1.6$ & $-14.3$ & 3/13 \\
 & Node classification (5) & $-2.3$ & $-18.6$ & 1/5 \\
 & Link prediction (5) & $-1.6$ & $-16.4$ & 1/5 \\
 & Cross-modal retrieval (4) & $-0.7$ & $-6.2$ & 1/4 \\
\midrule
\rowcolor{magL}Relation-Bank Partition & Macro (13) & $-0.1$ & $-0.2$ & 0/13 \\
 & Node classification (5) & $-0.2$ & $-0.6$ & 0/5 \\
 & Link prediction (5) & $-0.1$ & $+0.3$ & 0/5 \\
 & Cross-modal retrieval (4) & $+0.1$ & $-0.4$ & 0/4 \\
\bottomrule[1.2pt]
\end{tabular}
\par\vspace{2pt}
{\scriptsize Macro rows match Table~\ref{tab:q4-stability-summary}. Retention averages sources with positive-gain mass within each task. Build rate and flips use the 13 parent sources. Appendix~\ref{sec:appendix_corruption} defines each perturbation.}
\end{table}

\end{document}